\documentclass[11pt, a4paper, logo, copyright, nonumbering]{wechat}
\usepackage[numbers, square, sort&compress]{natbib}
\usepackage{dblfloatfix}
\usepackage{ulem}
\usepackage{caption}
\definecolor{citecolor}{HTML}{1976D2}
\hypersetup{
    colorlinks=true,
    linkcolor=black,
    filecolor=magenta,
    urlcolor=blue!50!black,
    citecolor=citecolor,
}
\usepackage{dramatist}
\usepackage{xspace}
\usepackage{pifont}
\usepackage{multirow}
\usepackage{tcolorbox}
\usepackage{xltabular}
\usepackage{longtable}
\usepackage{hyperref}
\usepackage{diagbox}
\usepackage{makecell}
\usepackage{amsfonts}
\usepackage{amsmath}
\usepackage{amssymb}
\usepackage{lineno}

\usepackage[bottom]{footmisc}

\usepackage{CJKutf8}
\usepackage{subfigure}
\usepackage{setspace}
\usepackage{subcaption} 

\usepackage{titlesec}

\usepackage{tcolorbox}
\usepackage{fontawesome5}
\usepackage{enumitem}

\usepackage{booktabs}
\usepackage{multirow}
\usepackage[table,xcdraw]{xcolor}
\usepackage{graphicx}

\usepackage{algorithm}
\usepackage{algpseudocode}

\tcbset{
    contributionbox/.style={
        colback=blue!3,
        colframe=blue!50!black,
        boxrule=0.5pt,
        arc=3pt,
        left=6pt,
        right=6pt,
        top=6pt,
        bottom=6pt,
        fonttitle=\bfseries
    }
}

\titlespacing*{\paragraph}
  {0pt}                   
  {0.5ex plus 1ex minus .2ex} 
  {1em}            

\makeatletter
\def\@BTrule[#1]{%
  \ifx\longtable\undefined
    \let\@BTswitch\@BTnormal
  \else\ifx\hline\LT@hline
    \nobreak
    \let\@BTswitch\@BLTrule
  \else
     \let\@BTswitch\@BTnormal
  \fi\fi
  \global\@thisrulewidth=#1\relax
  \ifnum\@thisruleclass=\tw@\vskip\@aboverulesep\else
  \ifnum\@lastruleclass=\z@\vskip\@aboverulesep\else
  \ifnum\@lastruleclass=\@ne\vskip\doublerulesep\fi\fi\fi
  \@BTswitch}
\makeatother

\addto\extrasenglish{
}

 {\begin{list}{}%
         {\setlength{\leftmargin}{#1}}%
         \item[]%
 }
 {\end{list}}

\reportnumber{001} 

\renewcommand{\today}{}

\title{\centering Hidden Decoding at Scale: Latent Computation Scaling for Large Language Models}

\author{
    {\large \bfseries WeChat AI Team}
}

\renewcommand{\phi}{\varphi}

\renewcommand{\epsilon}{\varepsilon}
\renewcommand{\imath}{\mathrm{i}}

\newlength{\restsubwidth}
\newlength{\restsubheight}
\newlength{\restsubmoreheight}
\newcommand{\rest}[2]{%
        \settowidth{\restsubwidth}{\ensuremath{#2}}
        \settoheight{\restsubheight}{\ensuremath{{}_{#2}}}
        \ensuremath{{#1\hskip 0.5pt}_{\vrule\kern2pt\parbox[b][%
        4pt][b]{\the\restsubwidth}{%
                        \ensuremath{{}_{#2}}}}}
        }

\begin{abstract}
Scaling Large Language Models (LLMs) has been driven mainly by enlarging the Transformer backbone, but for an already-strong model this requires another round of costly pretraining. We study whether an existing backbone can keep improving by allocating more computation to each token while leaving the Transformer backbone fixed. Depth-recurrent (looped) Transformers pursue this goal but are hard to scale, because looped computation does not fit naturally with the pipeline parallelism used to train the largest models. We add computation along the sequence-length dimension, where the extra computation is simply a longer input and stays compatible with standard large-model training.
We propose \textbf{\texttt{Hidden Decoding}}, a sequence-length scaling method applied during continued pretraining (CPT). It expands each token into $n$ streams with independent embedding tables and keeps the intermediate streams' key--value cache as context, so each token performs more internal computation without adding or widening Transformer layers. To keep this affordable at scale, we introduce \textit{Stream-Factorized Attention}, in which most layers attend only within each stream and only a few layers mix across streams, reducing the attention cost from quadratic to roughly linear in $n$. Experiments support two scaling results. At frontier scale, we train WeLM-HD4-80B and WeLM-HD4-617B at $n{=}4$ and improve their matched non-HD baselines, making Hidden Decoding the first demonstrated sequence-length scaling method at the 100B+ MoE scale. Across expansion factors, the gains grow as $n$ increases, showing that sequence-length expansion is a practical fixed-backbone scaling path for frontier-scale LLMs.
\end{abstract}

\begin{document}
\begin{CJK*}{UTF8}{gbsn}

\maketitle

\enlargethispage{1cm}

\newcommand\blfootnote[1]{%
  \begingroup
  \renewcommand\thefootnote{}\footnote{#1}%
  \addtocounter{footnote}{-1}%
  \endgroup
}

\begin{figure}[h]
    \centering
    \includegraphics[width=\textwidth]{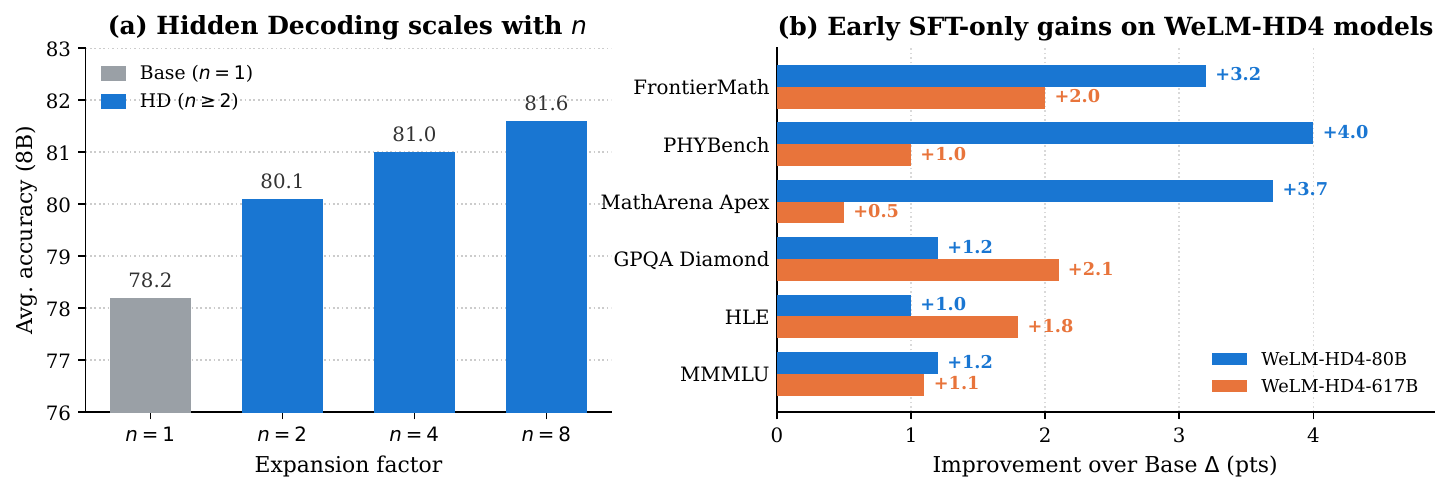}
    \caption{
    \textbf{Hidden Decoding scales and improves frontier models.} \textbf{(a)} On the dense Qwen3-8B-Base, average accuracy grows steadily with the expansion factor $n$. \textbf{(b)} Improvement $\Delta$ (in points) of WeLM-HD4-80B and WeLM-HD4-617B over their matched non-HD counterparts, WeLM-80B and WeLM-617B, after the same early SFT-only post-training.
    }
    \label{fig:teaser}
\end{figure}

\section{Introduction}

Scaling has been a primary source of progress in Large Language Models (LLMs). Increasing model size, training data, and training compute has consistently led to stronger models \citep{kaplan2020scaling,hoffmann2022training}. However, for already strong foundation models, further scaling the Transformer backbone is costly: it often requires another round of large-scale pretraining and increases both training and serving costs. This motivates the setting of this paper: improving an existing backbone by giving each token more internal computation while keeping the Transformer backbone fixed.

Recent reasoning models and test-time scaling studies show that, even with fixed parameters, spending more computation per problem can improve accuracy \citep{openai2024reasoning,deepseek2025r1,muennighoff2025s1,snell2024scaling}. This motivates a goal that recent latent-reasoning work has started to pursue: moving part of the extra thinking computation from visible tokens into the model's internal computation.

A prominent direction toward this goal is recurrent-depth or looped Transformers, which add computation by reusing the same Transformer blocks several times on each token \citep{geiping2025latent_recurrent,zhu2025ouro,saunshi2025reasoninglatentthoughtspower,park2026loopus}. This is a direct way to increase per-token computation with a fixed set of Transformer weights. However, it is hard to scale. Training the largest MoE models relies on pipeline parallelism \citep{huang2019gpipe,narayanan2021efficient,deepseekai2025deepseekv3technicalreport,kimiteam2026kimik2openagentic}, which assumes each input passes through the model's stages only once. A looped model breaks this assumption, because it feeds the same hidden states back through the same stages on every iteration, which stalls the pipeline and leaves GPUs idle. Consistent with this, looped models have stayed small: the largest is the 40B dense LoopCoder, trained without pipeline parallelism \citep{yang2026iquestcoderv1technicalreport}, while other looped language models remain at a few billion parameters \citep{geiping2025latent_recurrent,zhu2025ouro}. Looping therefore still lacks a practical way to scale to very large models while keeping GPUs efficiently utilized.

A key observation is that sequence-dimension computation fits large-scale training better than depth reuse. Along the sequence dimension, the expansion simply amounts to feeding a longer input, which is naturally compatible with the standard optimizations used for large-model training. The general paradigm is to expand each input token into several streams along the sequence, process the expanded sequence in a single forward pass, and apply the next-token loss only to the final stream, so the intermediate streams add per-token computation without adding new Transformer layers or widening existing ones. One instantiation is the Parallel Hidden Decoding Transformer (PHD) \citep{wu2025efficientpretraininglengthscaling}, which repeats each token but, to keep inference cheap, makes the repeated-token KV transient: intermediate streams can help within the current token, but they are not retained as independent KV context for later tokens. The capability-scaling setting therefore needs persistent intermediate-stream states that remain available across positions during CPT.

In this work, we propose a sequence-length scaling method named \textbf{\texttt{Hidden Decoding}}, which improves the capability of frontier-scale LLMs through continued pretraining (CPT). Two design choices make the expanded streams useful for capability scaling. To give different streams distinct initial states, we replicate the vocabulary embedding table into $n$ per-stream tables, expanding each token $x_i$ into streams $(E_1(x_i), E_2(x_i), \ldots, E_n(x_i))$; through CPT, these tables learn diverse initial representations of the same token. To let intermediate computation persist across positions, we keep the KV of the intermediate streams, so the computation accumulated across streams stays available as context for later tokens.

Retaining the KV of all streams makes intermediate computation available to later tokens, but it also makes the attention cost grow as $O(n^2L^2)$, which becomes infeasible to train for large models with long sequences. We therefore introduce \textit{Stream-Factorized Attention}. The key idea is to limit cross-stream attention to a subset of layers: most layers attend only within each stream, and the streams exchange information only at the remaining layers, which follow the base model's attention pattern (sliding-window, or full when the base model uses full attention). Because dense attention over the full $nL$ sequence is thus avoided at most layers, the added attention cost stays roughly linear rather than quadratic in $n$. This efficiency is what makes Hidden Decoding practical to train at the scale of MoE models with over 100B parameters, a regime not reached by prior looped or length-scaling methods.

Experiments directly test the two scaling claims. To test whether Hidden Decoding scales to the 100B+ MoE regime, we train WeLM-HD4-80B and WeLM-HD4-617B at $n{=}4$. WeLM-HD4-80B improves all nine shared benchmarks over WeLM-80B, with large gains on SciCode ($45.8\to50.0$) and PHYBench ($69.8\to73.8$). WeLM-HD4-617B also improves all nine shared benchmarks over WeLM-617B, including GPQA Diamond ($89.1\to91.2$), HLE ($33.6\to35.4$), and FrontierMath ($49.0\to51.0$). These results use an early SFT-only post-training recipe: each matched pair of non-HD and HD models uses the same SFT training recipe, with a short supervised fine-tuning schedule and no reinforcement learning. We use them as controlled comparisons; mature WeLM release scores are outside the scope of this paper. To test whether the expansion factor itself provides scaling, we increase $n$ from $2$ to $8$ in the 80B progressive-expansion study. MMLU improves monotonically ($85.0\to86.7\to87.5$), and Pile-test BPB falls from $0.386$ to $0.378$. Together, these results establish sequence-length scaling as a practical fixed-backbone scaling path for frontier-scale LLMs.

\begin{tcolorbox}[contributionbox, title={\faLightbulb~Main Contributions}]
\begin{itemize}[leftmargin=*, itemsep=6pt, label=\textcolor{blue!70!black}{\faCheckCircle}]
    \item \textbf{Frontier-Scale Fixed-Backbone Scaling.} To our knowledge, we are the first to demonstrate sequence-length scaling at the 100B+ MoE scale through continued pretraining (CPT). This yields two key models, WeLM-HD4-80B and WeLM-HD4-617B, in a regime not reached by prior depth-recurrent (looped) or length-scaling approaches.
    \item \textbf{Efficient Sequence-Length Expansion.} To make the expanded sequence trainable at frontier scale, we introduce \textit{Stream-Factorized Attention}: most layers attend only within each stream, while a small subset mixes streams. This keeps the attention cost near-linear in $n$; on WeLM-HD4-80B and WeLM-HD4-617B, the $4\times$ expanded sequence costs only $5.1\times$ and $4.4\times$ per batch, respectively, far below the dense-attention $16\times$ baseline.
    \item \textbf{Expansion-Factor Scaling.} We show that increasing the expansion factor improves language-modeling loss and downstream accuracy, validating $n$ as a practical scaling knob for a fixed Transformer backbone.
\end{itemize}
\end{tcolorbox}

\section{Method}
\label{sec:method}

\begin{figure}[!t]
    \centering
    \includegraphics[width=\linewidth]{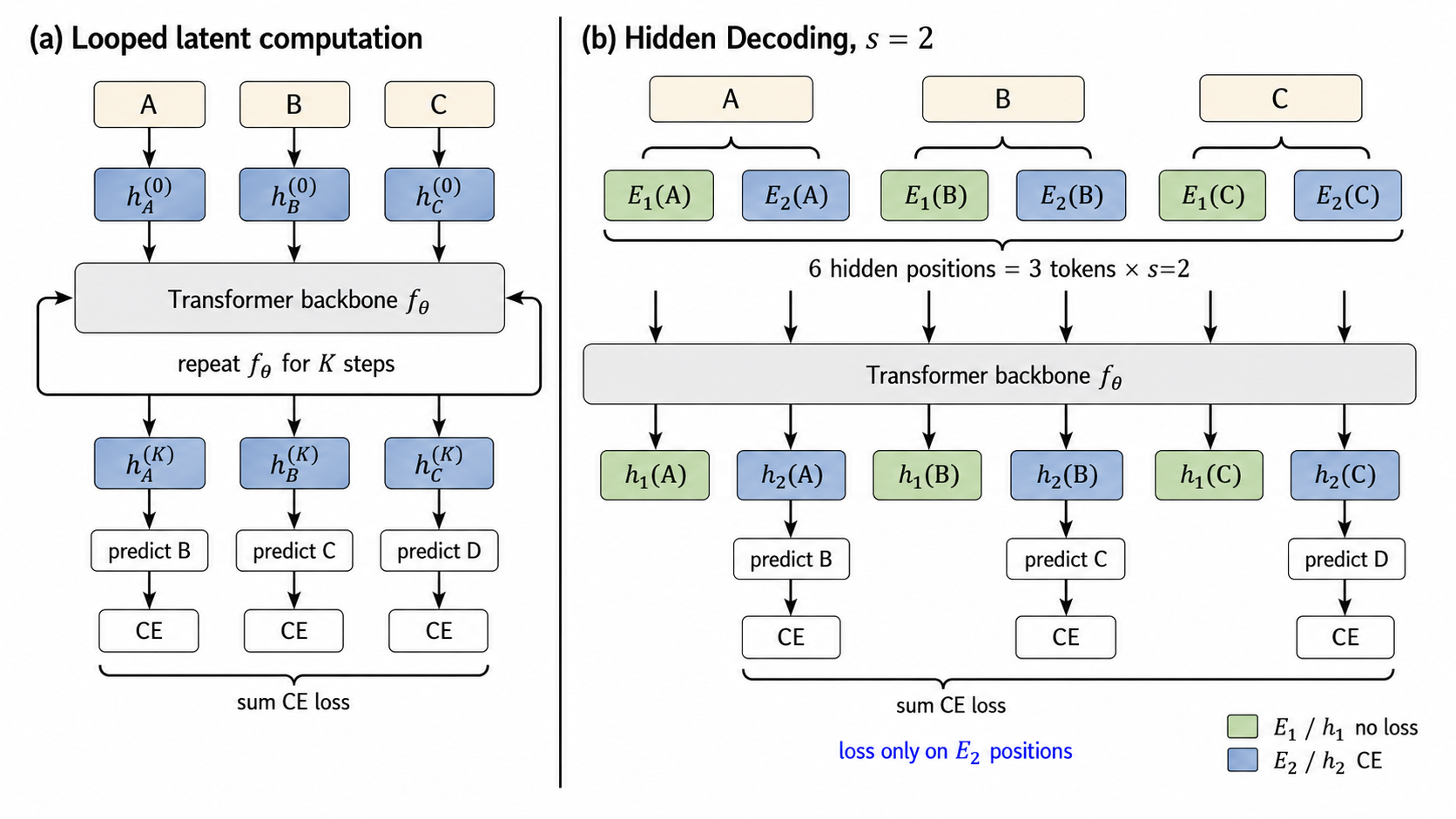}
    \caption{\textbf{Looped latent computation vs.\ Hidden Decoding.} \textbf{(a)} Depth-recurrent (looped) computation reuses the same backbone $f_\theta$ for $K$ steps on each token's hidden state before predicting the next token. \textbf{(b)} Hidden Decoding expands each token into $n$ streams with independent embedding tables $E_1,\dots,E_n$, interleaving them into a length-$nL$ sequence that is processed in a single pass through the same backbone. The next-token cross-entropy is applied only at the final-stream positions ($E_n/h_n$); the earlier streams ($E_1,\dots,E_{n-1}$) receive no loss and act as latent computation states. The panel shows a two-stream example.}
    \label{fig:method}
\end{figure}

\texttt{Hidden Decoding} is a sequence-length scaling method that increases the computation each token receives without enlarging the Transformer backbone. Each input token is represented in $n$ parallel streams inside the Transformer, so a length-$L$ sequence is processed as a length-$nL$ sequence in a single forward pass, giving each token $n$ internal computation steps before its prediction. Figure~\ref{fig:method} contrasts Hidden Decoding with depth-recurrent (looped) computation: Hidden Decoding places the extra computation along the sequence dimension, while looped computation reuses the backbone along the depth dimension. This section formalizes the multi-stream expansion and training objective (\S\ref{subsec:expansion}), introduces \textit{Stream-Factorized Attention} to keep the expansion affordable at scale (\S\ref{subsec:sfa}), and describes how to grow the expansion factor progressively from a converged checkpoint (\S\ref{subsec:progressive}).

\paragraph{Setup and notation.}
A standard Transformer language model maps each token $x_i$ through a single vocabulary embedding table $E$ to form the input sequence $(E(x_1), \dots, E(x_L))$; it adds positions with Rotary Positional Embeddings (RoPE)~\citep{su2024roformer}, applies causal attention, and predicts the next token from each position's hidden state through a shared head $g_\theta(\cdot)$. Hidden Decoding leaves this path unchanged and changes only how each token is embedded into the sequence.

\subsection{Multi-Stream Token Expansion}
\label{subsec:expansion}

The core construction of Hidden Decoding is to replace the single embedding table of a standard model with $n$ independent embedding tables, which we call \emph{streams}, and to interleave their outputs into one longer input sequence.

\paragraph{Expanded sequence.}
Given an input sequence $X = (x_1, \dots, x_L)$ and $n$ embedding tables $E_1, \dots, E_n$, we form an expanded sequence $S$ of length $nL$ by placing the stream-$k$ representation of token $x_i$ at physical position $t = (i-1)n + k$:
\begin{equation}
    S_t = E_k(x_i), \qquad 1 \le i \le L, \quad 1 \le k \le n .
    \label{eq:expansion}
\end{equation}
For $n=2$, this yields $S = \big(E_1(x_1), E_2(x_1), E_1(x_2), E_2(x_2), \dots\big)$. The expanded sequence is fed to the same Transformer under standard causal attention, with contiguous RoPE positions $0, 1, \dots, nL-1$. The expansion factor $n$ directly controls how much extra per-token computation is added.

\paragraph{Training objective.}
Only the final stream of each token, $E_n(x_i)$, is supervised: from its hidden state we predict the next token $x_{i+1}$ through the shared language-modeling head $g_\theta$, while the first $n-1$ streams receive no direct loss. Letting $h_t$ denote the hidden state at physical position $t$, the objective is the next-token cross-entropy applied only at the final-stream positions $t = in$:
\begin{equation}
    \mathcal{L}(\theta) = - \sum_{i=1}^{L} \log g_\theta\!\big(x_{i+1} \mid h_{in}\big).
    \label{eq:hd_loss}
\end{equation}
Because attention is causal, the final stream $E_n(x_i)$ attends to all earlier streams of the same token and of all preceding tokens. The first $n-1$ streams therefore act as intermediate computation states that progressively refine the representation before the final, prediction-bearing stream. Supervising only the final stream and giving each stream its own embedding table give the intermediate streams this latent-computation role; supervising all streams, or summing their outputs before prediction, degrades performance in the supervision-design ablation in \S\ref{subsec:sup_design}.

\subsection{Stream-Factorized Attention}
\label{subsec:sfa}

Running dense attention over the $nL$ expanded positions costs $O(n^2L^2)$, so the training cost grows sharply for large LLMs with long sequences. \textit{Stream-Factorized Attention} reduces this by making most layers attend only \emph{within} a stream and confining \emph{across}-stream mixing to a subset of layers.

\paragraph{Stream and token index.}
For a position $t \in \{1,\dots,nL\}$ in the expanded sequence, let its token and stream indices be
\begin{equation}
    i(t) = \left\lceil t/n \right\rceil, \qquad s(t) = \big((t-1)\bmod n\big) + 1,
    \label{eq:indices}
\end{equation}
so that $S_t = E_{s(t)}(x_{i(t)})$. Consecutive blocks of $n$ positions correspond to consecutive tokens, with the $n$ streams ordered inside each block.

\begin{figure}[!t]
    \centering
    \includegraphics[width=\linewidth]{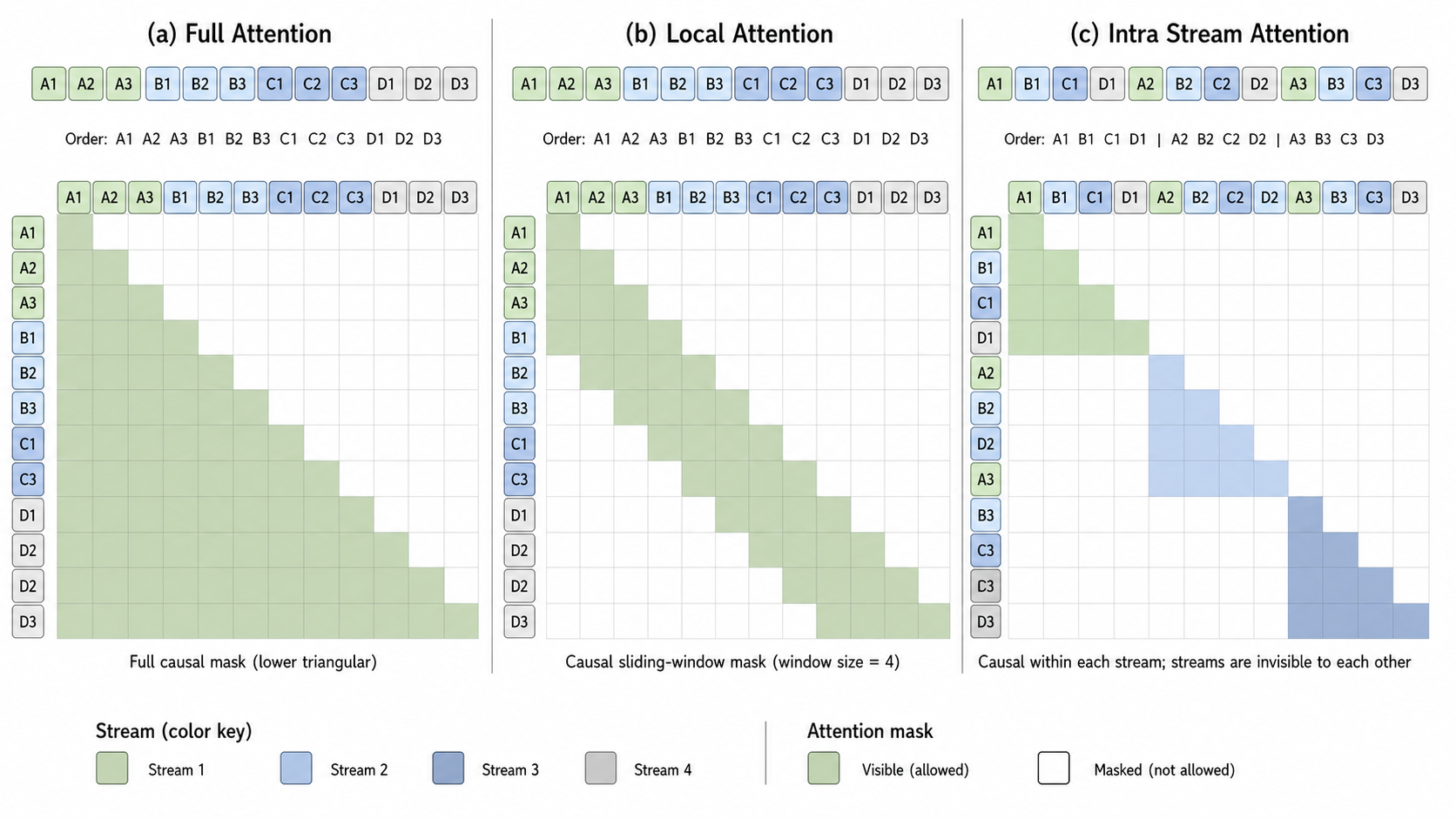}
    \caption{\textbf{Attention masks in Stream-Factorized Attention.} Each panel shows which positions a query (row) may attend to (colored) in the expanded sequence. \textbf{(a)} A full cross-stream layer uses the standard causal (lower-triangular) mask. \textbf{(b)} A local cross-stream layer uses a causal sliding-window mask. \textbf{(c)} An intra-stream layer is causal within each stream, so the streams are invisible to one another. Stream-Factorized Attention makes most layers intra-stream (c) and enables cross-stream mixing (a or b) at only a few layers.}
    \label{fig:masks}
\end{figure}

\paragraph{Intra-stream and cross-stream layers.}
Using the indices in Eq.~\ref{eq:indices}, an \emph{intra-stream} layer restricts attention to earlier positions of the same stream,
\begin{equation}
    M^{\text{intra}}_{t,t'} = \mathbf{1}\!\left[\, t' \le t \ \wedge\ s(t') = s(t) \,\right],
    \label{eq:mask_intra}
\end{equation}
which costs $O(nL^2)$ instead of the $O(n^2L^2)$ of a dense layer; a \emph{cross-stream} layer also attends across streams, following the model's usual attention pattern. Figure~\ref{fig:masks} illustrates the three resulting mask types.

\paragraph{Design.}
Most layers are intra-stream, and cross-stream attention is enabled at only a subset of layers. We add no new attention layers for this: we reuse the base model's own attention layers as the cross-stream layers. When the base model contains sliding-window attention, those layers serve as the cross-stream layers, so cross-stream attention is local; otherwise the cross-stream layers are full-attention layers. Because most layers are intra-stream, the added attention cost stays close to linear in $n$ rather than quadratic (\S\ref{sec:cost}); the number and placement of cross-stream layers is a design choice we study in \S\ref{sec:experiments}. This layout also keeps CPT closer to the checkpoint before HD expansion: most layers preserve the base model's single-stream causal path, while only a few layers introduce cross-stream perturbations; Appendix~\ref{app:cpt_startup_loss} gives a separate 617B-scale startup-loss check for this effect.

\subsection{Progressive Expansion}
\label{subsec:progressive}

A large expansion factor $n$ can be trained directly, but progressively growing it provides a better initialization for the many embedding streams and saves compute. We therefore grow the factor progressively---$1 \!\to\! 2 \!\to\! 4 \!\to\! 8$---initializing each factor from the converged checkpoint of the previous one. Concretely, we use \emph{Cyclic Replication Initialization}: when doubling from $n$ to $2n$, the new streams copy the existing tables,
\begin{equation}
    E_{n+k} \leftarrow E_k, \qquad 1 \le k \le n,
    \label{eq:cyclic_init}
\end{equation}
so that the expanded model initially reproduces the behavior of the smaller one and then adapts smoothly. For example, $n{=}2 \!\to\! 4$ initializes the tables as $(E_1, E_2, E_1, E_2)$, and $n{=}4 \!\to\! 8$ as $(E_1, E_2, E_3, E_4, E_1, E_2, E_3, E_4)$. In practice, we introduce successive expansions at scheduled token counts during continued pretraining, letting each factor converge before the next expansion.

\section{Computational Cost}
\label{sec:cost}

Hidden Decoding adds computation to every token, so its training and serving costs must be measured. We separate two kinds of cost: the cost inherent to the method, which holds on any backbone, and an extra saving specific to our WeLM backbone. We report the training measurements in \S\ref{subsec:main_results} and the serving measurements in \S\ref{subsec:throughput}.

\subsection{Training Cost}
\label{subsec:kvmirror}

Expanding each token into $n$ streams turns a length-$L$ sequence into length $nL$. Non-attention computation scales roughly with the number of positions, but dense attention over the expanded sequence would scale as $O(n^2L^2)$, or $n^2\times$ the attention cost of the unexpanded sequence. Stream-Factorized Attention (\S\ref{subsec:sfa}) avoids this dense-attention blow-up by keeping most layers intra-stream, so the attention cost stays close to linear in $n$. This near-linear attention design is inherent to the method and holds on any backbone.

Our WeLM backbones allow one further, architecture-specific saving through their \emph{KV-mirror} design~\citep{welmblog}. In a standard layer, the keys and values come from that layer's own input hidden states; under KV-mirror, each later layer instead computes them from an earlier layer's hidden states, using its own projection weights. The layers are paired in a U-shape, with the last third mirroring the first third. Because a mirrored layer's keys and values depend only on the early layer's hidden states, our training framework computes them at the early layers, and the mirrored layers reuse them directly.

KV-mirror lets Hidden Decoding skip work in mirrored layers. Only the final stream is supervised (\S\ref{subsec:expansion}), so the intermediate streams matter only through the keys and values they add as context. In the mirrored layers this context is already fixed by the early layers, and the intermediate streams carry no loss, so these layers need not process the intermediate streams at all. \textbf{We therefore run only the final stream through the mirrored layers, while the early layers still process all $n$ streams.} In our 80B and 617B models the last third of the layers are mirrored, so the intermediate streams are skipped across that final third. This saving is specific to WeLM; it comes on top of the near-linear cost that Stream-Factorized Attention already provides on any backbone. As an implementation check, this KV-mirror optimization reduces per-batch time from $15\,\mathrm{s}$ to $12\,\mathrm{s}$ in an 80B 32k training setting, a $20\%$ reduction ($1.25\times$ speedup). We use this number to quantify the WeLM-specific training-system saving used by WeLM-HD4-80B and WeLM-HD4-617B; the main cost comparison in \S\ref{subsec:main_results} reports end-to-end HD training time under the long-context settings. The net effect is that training cost grows near-linearly with the number of streams $n$. We measure this at both scales in \S\ref{subsec:main_results} (Figure~\ref{fig:exp}a).

\subsection{Inference Cost}
\label{subsec:inference}

A key advantage of Hidden Decoding over looped models is that its extra computation runs in parallel. A looped model adds computation by repeating the backbone step by step, and each step waits for the previous one, so the added cost cannot be hidden. Hidden Decoding spreads its computation across the $n$ streams, which are processed together in a single forward pass. At large batch sizes, decoding is compute-bound, so this extra computation lowers throughput. At small batch sizes, decoding is bound by memory bandwidth and leaves much of the compute idle; the parallel streams can use that idle compute, so the added cost stays small. The one remaining overhead in this regime is the larger KV cache, which Stream-Factorized Attention keeps modest by reading the full expanded KV at only a few layers. \textbf{The throughput penalty is therefore smallest in the latency-sensitive, short-input/small-batch regime and grows with larger batches or longer inputs}, as we measure in \S\ref{subsec:throughput}.

For WeLM models, KV-mirror can also reduce decoding-side computation by skipping mirrored-layer work for intermediate streams. The serving benchmark in \S\ref{subsec:throughput} uses the current benchmark implementation without this additional KV-mirror decoding optimization, because integrating it into the decoding kernel requires separate engineering. The reported throughput therefore does not include this WeLM-specific speedup.

\section{Experiments}
\label{sec:experiments}

We use the experiments to test three parts of Hidden Decoding (HD). The frontier-scale MoE study checks whether HD improves the largest models we deploy, producing WeLM-HD4-80B and WeLM-HD4-617B at acceptable training cost (\S\ref{subsec:main_results}). The expansion-factor study checks whether adding more streams behaves like a scaling axis (\S\ref{subsec:progressive_exp}). The ablations and throughput measurement check which attention choices and serving costs matter in practice (\S\ref{subsec:ablations}, \S\ref{subsec:throughput}).

\definecolor{highlightblue}{RGB}{235, 245, 255}
\definecolor{headergray}{RGB}{242, 242, 242}

\subsection{Setup}
\label{subsec:exp_setup}

\paragraph{Models.}
We apply HD on top of strong pre-trained checkpoints through continued pretraining (CPT), without enlarging the Transformer backbone. Our main HD models are WeLM-HD4-80B and WeLM-HD4-617B, obtained by expanding WeLM MoE checkpoints at 80B (3B activated) and 617B (23B activated) total parameters to $n{=}4$. Their matched non-HD counterparts are WeLM-80B and WeLM-617B. When reporting base-model evaluations before post-training, we append \texttt{-Base}, giving WeLM-80B-Base vs.\ WeLM-HD4-80B-Base and WeLM-617B-Base vs.\ WeLM-HD4-617B-Base. We use smaller 21B (0.7B activated) and 6B (0.6B activated) MoE models for ablations, and the dense Qwen3-8B-Base~\citep{qwen3}\begin{NoHyper}\footnote{\url{https://github.com/Tencent/Sequential-Hidden-Decoding}.}\footnote{\url{https://huggingface.co/collections/tencent/sequential-hidden-decoding}.}\end{NoHyper} for progressive expansion and probes.

Only the embedding tables grow with the expansion factor $n$, while the active Transformer parameters per token stay unchanged. For example, expanding to $n{=}4$ increases the stored embedding parameters from $26.9$B to $107.4$B for WeLM-HD4-617B and from $1.2$B to $3.1$B on Qwen3-8B-Base (reaching $5.6$B at $n{=}8$). These tables are sparse lookups: each position reads a single embedding row, and the tables do not enter the attention or feed-forward matrix multiplications that set the compute cost. We therefore report the embedding growth separately and compare matched non-HD and HD models as a fixed-Transformer-backbone, fixed-active-parameter test.

\paragraph{Configurations.}
The large MoE models are expanded to $n{=}4$. The dense model is expanded progressively to $n\in\{2,4,8\}$ for the expansion-factor study. Following \S\ref{subsec:sfa}, most layers are intra-stream and cross-stream mixing is placed on a subset of layers. For the 80B model (49 layers), we use 6 full, 20 intra-stream, and 23 sliding-window layers. For the 617B model (94 layers), we use 25 full and 69 intra-stream layers. The base models' native context length is 256k for 80B and 32k for 617B. RoPE and context-extension settings are held fixed within each matched comparison between non-HD and HD models, including the shared early SFT recipe in Appendix~\ref{app:sft_recipe}. Appendix~\ref{app:model_config} gives the backbone and HD configuration details for the two main WeLM models.

\paragraph{Continued pretraining.}
We introduce Hidden Decoding during continued pretraining. We start from a baseline checkpoint and continue training with the expansion enabled, using the same data and schedule as the baseline except for the 617B long-context-stage difference in Table~\ref{tab:cpt}. The expansion stays on from that point onward: it carries from the 32k continuation stage into a later 256k long-context stage, and then through the early SFT-only post-training recipe. The non-HD baseline is trained in the same way with the expansion turned off, so the two runs differ only in Hidden Decoding. For the 80B comparison, the two runs see identical data at every stage. For the 617B comparison, the HD long-context stage uses fewer tokens than the non-HD model, so its gains are a conservative lower bound.

Table~\ref{tab:cpt} reports the training-token budget behind this comparison.

\begin{table}[H]
\centering
\small
\setlength{\tabcolsep}{4pt}
\caption{\textbf{Continued-pretraining token budgets for non-HD base paths and HD4 continuation windows.} Non-HD rows count the full base-model path. HD4 rows count only the training after the HD start point.}
\label{tab:cpt}
\begin{tabular}{@{}llccccc@{}}
\toprule
\textbf{Scale} & \textbf{Run} & \makecell{\textbf{HD start}\\\textbf{tokens}} & \makecell{\textbf{8k}\\\textbf{pretrain}} & \makecell{\textbf{32k}\\\textbf{continuation}} & \makecell{\textbf{256k}\\\textbf{continuation}} & \textbf{Total} \\
\midrule
80B  & WeLM-80B-Base        & -- & 17.81T & 2.01T & 0.57T & 20.39T \\
80B  & \cellcolor{highlightblue}WeLM-HD4-80B-Base  & \cellcolor{highlightblue}19.32T & \cellcolor{highlightblue}0 & \cellcolor{highlightblue}0.50T & \cellcolor{highlightblue}0.57T & \cellcolor{highlightblue}1.07T \\
617B & WeLM-617B-Base       & -- & 14.25T & 2.20T & 0.61T & 17.06T \\
617B & \cellcolor{highlightblue}WeLM-HD4-617B-Base & \cellcolor{highlightblue}15.86T & \cellcolor{highlightblue}0 & \cellcolor{highlightblue}0.59T & \cellcolor{highlightblue}0.30T & \cellcolor{highlightblue}0.90T \\
\bottomrule
\end{tabular}
\end{table}

For 80B, Hidden Decoding starts after $19.32$T tokens on the non-HD training path, and WeLM-HD4-80B-Base then trains on the same remaining $0.50$T 32k tokens and $0.57$T 256k tokens as the non-HD continuation. For 617B, Hidden Decoding starts after $15.86$T tokens on the non-HD training path. The HD4 run matches the remaining $0.59$T 32k continuation, but uses a shorter 256k continuation ($0.30$T vs.\ $0.61$T). Relative to the full non-HD base paths, the HD4-only token budgets are about $5.3\%$ at both scales, so the higher per-token cost is concentrated late in training.

\paragraph{Post-training scope.}
The post-training used for the early post-training tables is intentionally lightweight: it is an early supervised fine-tuning run and does not include reinforcement learning. We use this stage to test whether the gains from Hidden Decoding persist under the same downstream adaptation recipe for the non-HD and HD models. Appendix~\ref{app:sft_recipe} gives the SFT recipe. We report these scores as controlled early-SFT comparisons, with mature WeLM release scores left outside the scope of this paper.

\paragraph{Evaluation.}
We use two evaluation suites. For early post-training results (Tables~\ref{tab:main_moe} and~\ref{tab:main_80b}), we evaluate on recent hard benchmarks covering mathematics and science (HMMT and MathArena Apex~\citep{balunovic2025matharena}, IMO-AnswerBench~\citep{luong2025imobench}, FrontierMath~\citep{glazer2024frontiermath}, PolyMath~\citep{wang2025polymath}, GPQA~\citep{rein2024gpqa}, PHYBench~\citep{qiu2025phybench}, CritPt~\citep{zhu2025critpt}), knowledge and reasoning (HLE~\citep{phan2025hle}, MMMLU~\citep{openaimmmlu}, AA-OmniScience~\citep{jackson2025aaomniscience}, AA-LCR~\citep{aalcr}, ARC-AGI-2~\citep{chollet2025arcagi2}), code (SciCode~\citep{tian2024scicode}, LiveCodeBench~\citep{jain2024livecodebench}), instruction following (IFBench~\citep{pyatkin2025ifbench}), and agentic tasks (Terminal-Bench~2~\citep{terminalbench}, $\tau^2$-Bench~\citep{barres2025tau2}, GDPval~\citep{patwardhan2025gdpval}).

For scaling and ablation studies (\S\ref{subsec:progressive_exp}, \S\ref{subsec:ablations}) and pretraining-stage results (Appendix~\ref{app:pretrain}), we use a standard suite: MMLU~\citep{hendrycks2020measuring}, MMLU-Pro~\citep{wang2024mmlupro}, CMMLU~\citep{li2023cmmlu}, C-Eval~\citep{huang2023ceval}, SuperGPQA~\citep{du2025supergpqa}, ARC-C~\citep{clark2018think}, HellaSwag~\citep{zellers2019hellaswag}, BBH~\citep{suzgun2022bbh}, GSM8K~\citep{cobbe2021training}, MATH~\citep{hendrycks2021measuring}, SimpleQA~\citep{wei2024simpleqa}, Chinese SimpleQA~\citep{he2024chinesesimpleqa}, HumanEval+/MBPP+~\citep{liu2023code}, MultiPL-E~\citep{cassano2022multiple}, and CRUXEval~\citep{gu2024cruxeval}. Tables~\ref{tab:eval_config} and~\ref{tab:eval_config_sft} give the per-benchmark settings.

\subsection{Frontier-Scale MoE Results}
\label{subsec:main_results}

To test whether Hidden Decoding strengthens the largest models we deploy, we compare two forms of each WeLM MoE checkpoint that share everything except Hidden Decoding: WeLM-80B vs.\ WeLM-HD4-80B, and WeLM-617B vs.\ WeLM-HD4-617B. The HD models add Hidden Decoding during continued pretraining (Table~\ref{tab:cpt}); both forms are then adapted with the same early SFT-only post-training recipe, so the comparison isolates our method. Table~\ref{tab:main_moe} evaluates both scales on the same nine hard benchmarks.

\begin{table}[H]
    \centering
    \small
    \setlength{\tabcolsep}{6pt}
    \caption{\textbf{Early SFT$^{\clubsuit}$ results for WeLM-HD4-80B and WeLM-HD4-617B.} We compare each HD4 model with its matched non-HD counterpart: WeLM-80B vs.\ WeLM-HD4-80B and WeLM-617B vs.\ WeLM-HD4-617B. HD4 denotes Hidden Decoding with $n{=}4$; active Transformer parameters per token are unchanged ($3$B for 80B, $23$B for 617B). Both forms use the same early SFT-only post-training recipe, with a short SFT schedule and no RL. Kimi K2.6~\citep{moonshotai2026kimik26} is included as a mainstream frontier-model reference for the absolute score scale; its model card reports $1$T total parameters and $32$B activated parameters. $^{\ddagger}$HMMT is the weighted average over the Feb.\ 2025, Nov.\ 2025, and Feb.\ 2026 sets.}
    \label{tab:main_moe}
    \begin{tabular}{@{}l cc cc c@{}}
    \toprule
     & \multicolumn{2}{c}{\textbf{80B MoE}} & \multicolumn{2}{c}{\textbf{617B MoE}} & \multicolumn{1}{c}{\textbf{External}} \\
    \cmidrule(lr){2-3}\cmidrule(lr){4-5}\cmidrule(l){6-6}
    \textbf{Benchmark} & \makecell{\textbf{WeLM}\\\textbf{80B}} & \cellcolor{highlightblue}\makecell{\textbf{WeLM-HD4}\\\textbf{80B}} & \makecell{\textbf{WeLM}\\\textbf{617B}} & \cellcolor{highlightblue}\makecell{\textbf{WeLM-HD4}\\\textbf{617B}} & \makecell{\textbf{Kimi K2.6}\\\textbf{1T-A32B}} \\
    \midrule
    GPQA Diamond      & 87.6 & \cellcolor{highlightblue}88.8 & 89.1 & \cellcolor{highlightblue}91.2 & 90.4 \\
    HLE               & 27.4 & \cellcolor{highlightblue}28.4 & 33.6 & \cellcolor{highlightblue}35.4 & 36.9 \\
    MMMLU             & 84.4 & \cellcolor{highlightblue}85.6 & 86.4 & \cellcolor{highlightblue}87.5 & 88.0 \\
    FrontierMath$^{\star}$ & 45.8 & \cellcolor{highlightblue}49.0 & 49.0 & \cellcolor{highlightblue}51.0 & 53.2 \\
    PHYBench          & 69.8 & \cellcolor{highlightblue}73.8 & 75.3 & \cellcolor{highlightblue}76.3 & 74.0 \\
    MathArena Apex    & 16.4 & \cellcolor{highlightblue}20.1 & 24.2 & \cellcolor{highlightblue}24.7 & 23.8 \\
    HMMT$^{\ddagger}$ & 93.3 & \cellcolor{highlightblue}94.1 & 96.0 & \cellcolor{highlightblue}96.2 & 96.0 \\
    IMO-AnswerBench   & 85.0 & \cellcolor{highlightblue}85.3 & 87.5 & \cellcolor{highlightblue}88.5 & 91.5 \\
    SciCode           & 45.8 & \cellcolor{highlightblue}50.0 & 51.4 & \cellcolor{highlightblue}52.1 & 50.7 \\
    \bottomrule
    \end{tabular}
    \vspace{2pt}
    \begin{minipage}{0.95\linewidth}\raggedright
    \footnotesize $^{\clubsuit}$We use early SFT here to keep the non-HD and HD models matched in post-training data and configuration. A public report gives benchmark results for a newer non-HD WeLM MoE after larger-scale, higher-quality post-training: \url{https://welm.weixin.qq.com/en/posts/hidden_decoding_at_scale/\#conclusion-and-future-work}.\\
    \footnotesize $^{\star}$FrontierMath uses the public Tiers 1--4 sample problems: \url{https://epoch.ai/frontiermath/tiers-1-4/benchmark-problems}.\\
    \end{minipage}
    \end{table}

Hidden Decoding improves every benchmark at both scales. The largest gains appear on hard math and science tasks: WeLM-HD4-80B improves SciCode by $+4.2$, PHYBench by $+4.0$, and FrontierMath by $+3.2$; WeLM-HD4-617B improves GPQA by $+2.1$ and HLE by $+1.8$. We also include Kimi K2.6 (1T total, 32B activated) as a mainstream frontier-model reference for absolute score scale: WeLM-HD4-617B is higher on GPQA Diamond, PHYBench, MathArena Apex, HMMT, and SciCode, while Kimi K2.6 is higher on HLE, MMMLU, FrontierMath, and IMO-AnswerBench. On the broader 80B suite in Appendix~\ref{app:80b_extra}, the largest gains shift toward agentic and long-horizon tasks, including Terminal-Bench 2 ($44.9\!\to\!58.4$) and ARC-AGI-2 ($6.9\!\to\!11.6$). \textbf{These results show that Hidden Decoding can be scaled to frontier-size MoE models and improve them without enlarging the Transformer backbone.} This is the scaling regime where looped models have been hard to apply, because repeated depth passes fit poorly with the pipeline parallelism used to train large MoEs.

\begin{figure}[!t]
    \centering
    \includegraphics[width=\linewidth]{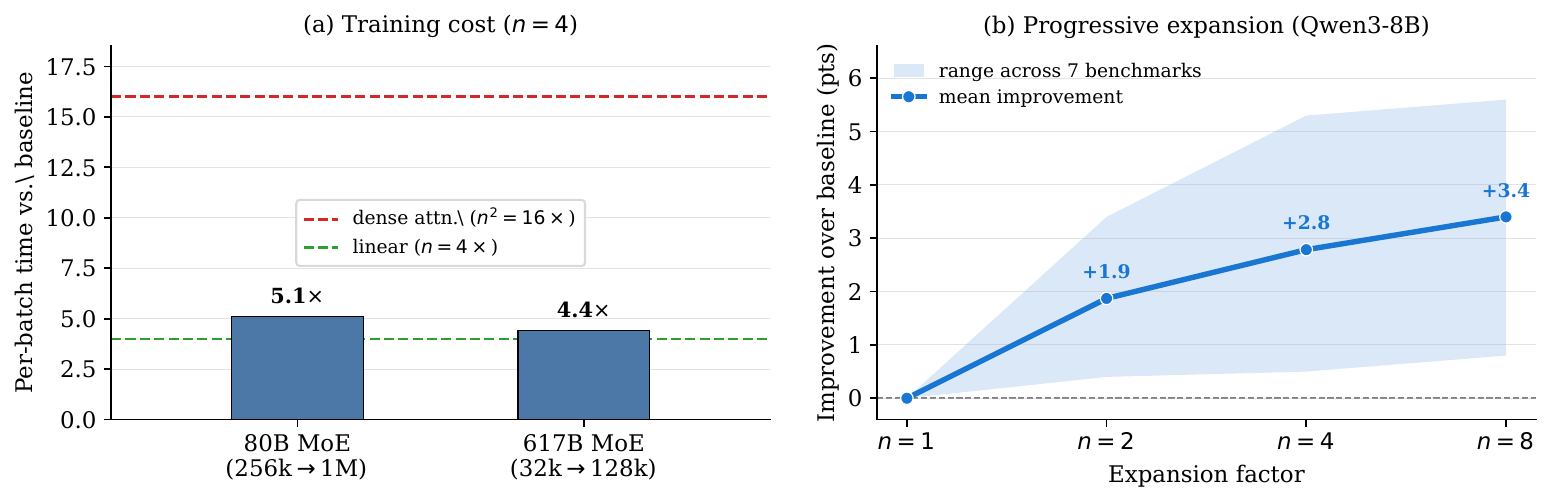}
    \caption{\textbf{(a) Training cost of HD ($n{=}4$):} per-batch time relative to the unexpanded baseline, staying near the linear ($n{=}4\times$) reference and far below dense attention ($n^2{=}16\times$). \textbf{(b) Progressive expansion on the dense Qwen3-8B-Base:} the mean improvement over the unexpanded baseline grows steadily with the expansion factor $n$; the shaded band shows the range across the seven benchmarks.}
    \label{fig:exp}
\end{figure}

\paragraph{Training cost measurement.}
To verify the training-cost analysis in \S\ref{sec:cost}, we measure per-batch training time on the WeLM backbones relative to the unexpanded baseline. Figure~\ref{fig:exp}(a) shows that a $4\times$ longer effective sequence costs $5.1\times$ on 80B ($256\mathrm{k}\!\rightarrow\!1\mathrm{M}$) and $4.4\times$ on 617B ($32\mathrm{k}\!\rightarrow\!128\mathrm{k}$). These costs stay close to the $4\times$ linear reference and far below the $16\times$ dense-attention cost. \textbf{This near-linear cost makes WeLM-HD4-617B trainable, while dense attention over the expanded sequence would be infeasible.} Using full attention on only a few layers is a deliberate trade-off; we measure its effect on accuracy with an ablation on a smaller model in \S\ref{subsec:ablations}.

\paragraph{External-model reference.}
To place the base-model strength of WeLM-HD4-617B-Base in context, we compare it with the base-model table reported in the Qwen3.5 report~\citep{qwen35blog}. Table~\ref{tab:external} keeps the overlapping benchmarks from that report and our base-model evaluation. We include both WeLM-617B-Base and WeLM-HD4-617B-Base, evaluated with our base-model protocol before instruction tuning or reinforcement learning. Because model families and evaluation harnesses still differ, the controlled evidence remains the matched comparison between non-HD and HD models under the same backbone, data, and training configuration. The WeLM baseline scores are obtained by evaluating the matched non-HD Base checkpoint with the same harness as the HD Base checkpoint.

\begin{table}[!t]
\centering
\small
\setlength{\tabcolsep}{4pt}
\caption{\textbf{Base-model reference.} The WeLM columns use our base-model protocol before instruction tuning or RL. External numbers are copied from the Qwen3.5 report's base-model table and provide absolute-score context; the matched WeLM columns give the controlled comparison between non-HD and HD models.}
\label{tab:external}
\begin{tabular*}{\linewidth}{@{\extracolsep{\fill}}l c >{\columncolor{highlightblue}}c ccc@{}}
\toprule
\textbf{Benchmark} & \makecell{\textbf{WeLM}\\\textbf{617B-Base}} & \makecell{\textbf{WeLM-HD4}\\\textbf{617B-Base}} & \makecell{DS-V3.2\\671B-A37B} & \makecell{K2\\1T-A32B} & \makecell{Qwen3.5\\397B-A17B} \\
\midrule
MMLU        & 89.85 & 90.22 & 88.11 & 87.38 & 88.61 \\
MMLU-Pro    & 72.26 & 73.92 & 62.82 & 67.64 & 76.01 \\
MMLU-Redux  & 89.81 & 90.94 & 87.29 & 86.65 & 89.09 \\
SuperGPQA   & 55.33 & 57.05 & 43.46 & 44.86 & 57.96 \\
C-Eval      & 91.90 & 92.50 & 90.48 & 91.82 & 91.82 \\
BBH         & 92.41 & 93.07 & 86.03 & 89.11 & 90.98 \\
KoRBench    & 52.08 & 54.88 & 54.00 & 53.84 & 54.08 \\
CRUX-Input  & 86.00 & 91.75 & 63.25 & 70.50 & 71.13 \\
CRUX-Output & 89.38 & 91.13 & 73.88 & 77.13 & 82.38 \\
\bottomrule
\end{tabular*}
\end{table}

\subsection{Expansion-Factor Scaling}
\label{subsec:progressive_exp}

To test whether the expansion factor is a usable scaling knob, we expand the dense Qwen3-8B-Base step by step to $n\in\{2,4,8\}$ while keeping the Transformer backbone fixed. Figure~\ref{fig:exp}(b) reports improvement over the unexpanded baseline. The mean improvement rises almost steadily with $n$, and every benchmark gains; the largest improvements reach $+5.6$ on HellaSwag and $+5.1$ on BBH and MATH.

The same trend appears at MoE scale. Expanding the 80B MoE raises MMLU from $85.1$ at $n{=}1$ to $87.5$ at $n{=}8$ and lowers Pile-test BPB from $0.386$ to $0.378$ (full per-factor results in Appendix~\ref{app:progressive}). \textbf{Within a fixed Transformer backbone, the expansion factor acts as a reliable scaling knob: increasing it improves accuracy and language-modeling loss.}

\begin{table}[!t]
    \centering
    \small
    \setlength{\tabcolsep}{6pt}
    \caption{\textbf{Attention-composition ablation (21B MoE).} We compare a no-expansion baseline with three HD variants that differ only in the number of full cross-stream layers. Header parentheses give the number of full-attention layers (out of $27$). SF with $4$ full layers nearly matches all-full, while SF with $1$ full layer still improves the baseline.}
    \label{tab:ablation_attn}
    \begin{tabular}{@{}l c >{\columncolor{highlightblue}}c >{\columncolor{highlightblue}}c >{\columncolor{highlightblue}}c@{}}
    \toprule
     & & \multicolumn{3}{c}{\textbf{Hidden Decoding}} \\
    \cmidrule(lr){3-5}
    \textbf{Benchmark} & Baseline & SF ($1$) & SF ($4$) & all-full ($27$) \\
    \midrule
    MMLU            & 74.1 & 75.7 & 76.4 & 76.3 \\
    MMLU-Pro        & 47.5 & 50.4 & 49.8 & 50.7 \\
    CMMLU           & 77.9 & 79.2 & 79.8 & 79.6 \\
    C-Eval          & 78.6 & 78.9 & 79.2 & 79.7 \\
    ARC-C           & 89.5 & 89.8 & 90.7 & 90.9 \\
    SuperGPQA       & 34.1 & 35.4 & 35.7 & 36.0 \\
    BBH             & 67.0 & 71.2 & 72.0 & 72.8 \\
    GSM8K           & 83.4 & 86.2 & 86.7 & 86.2 \\
    MATH            & 45.2 & 49.6 & 49.3 & 50.3 \\
    SimpleQA        & 3.7  & 4.0  & 3.6  & 4.1  \\
    AA-OmniScience  & 12.7 & 12.4 & 14.2 & 14.1 \\
    HumanEval+      & 37.4 & 37.6 & 39.2 & 40.5 \\
    MBPP+           & 55.2 & 56.1 & 57.8 & 59.1 \\
    \midrule
    \textbf{Average} & 54.33 & 55.88 & 56.49 & \textbf{56.95} \\
    \bottomrule
    \end{tabular}
    \end{table}

\subsection{Attention Composition Ablation}
\label{subsec:ablations}
To test how much full cross-stream mixing is required for accuracy, we ablate the attention composition inside Stream-Factorized Attention. Each layer mixes streams in one of three ways: \emph{full} (across all streams over the whole sequence), \emph{local} (across streams within a sliding window), or \emph{intra-stream} (no cross-stream mixing). Full mixing is the expensive case.

We run this ablation on the 21B MoE, which has $27$ layers and whose base model already interleaves full-attention and sliding-window layers. The reference is a standard no-expansion baseline. The three Hidden Decoding variants differ only in how many layers use full cross-stream mixing: $1$, $4$, or all $27$, with the remaining layers split between local and intra-stream mixing (exact layouts in Appendix~\ref{app:sfa}). Table~\ref{tab:ablation_attn} reports the evaluated benchmarks and their average.

Every Hidden Decoding model beats the no-expansion baseline on average, with gains from $+1.55$ to $+2.62$. The all-full model is best, but the cheaper $4$-full variant is close ($+2.16$), and the $1$-full variant still improves the baseline ($+1.55$). \textbf{A few full layers are enough: the remaining cross-stream mixing can be local or skipped, which keeps Stream-Factorized Attention cheap while preserving most of the accuracy gain.}

\subsection{Inference Throughput}
\label{subsec:throughput}
\begin{figure}[!t]
    \centering
    \includegraphics[width=0.88\linewidth]{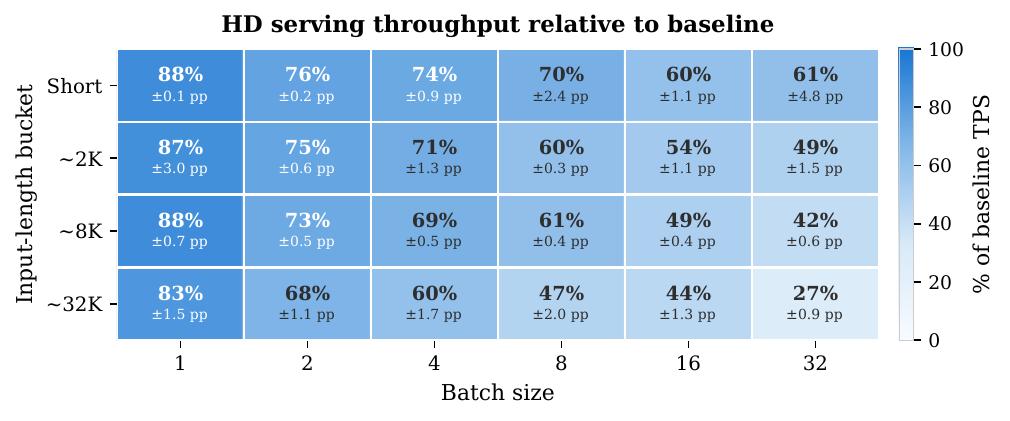}
\caption{\textbf{Serving throughput matrix for WeLM-HD4-80B.} Each cell reports mean decoding-only completion TPS for Hidden Decoding over five runs as a percentage of the matched WeLM-80B baseline with stream expansion disabled under the same input-length bucket and batch size; prefill latency is excluded. Small text shows the five-run sample standard deviation in percentage points. Both models run on the same $8$ H20 GPUs. The benchmark uses a $30$-second warmup, a $45$-second measurement window, and maximum output length $8192$.}
    \label{fig:throughput}
\end{figure}

To measure serving cost under a matched setting, we vary the two factors that most directly affect decoding throughput: input length and batch size. For WeLM-HD4-80B, we serve each request bucket twice on the same $8$ H20 GPUs: once with Hidden Decoding enabled, and once on the matched WeLM-80B baseline with stream expansion disabled. The benchmark reports decoding-only completion TPS after excluding prefill latency, with maximum output length $8192$, a $30$-second warmup, and a $45$-second measurement window. Each cell in Figure~\ref{fig:throughput} reports the mean Hidden Decoding completion throughput over five runs as a percentage of the matched baseline under the same input-length bucket and batch size; the small label gives the run-to-run standard deviation in percentage points.

The measurements identify a practical serving window. At batch size $1$, Hidden Decoding keeps $83$--$88\%$ of baseline throughput across all measured input buckets. For the short, 2k, and 8k buckets, it keeps $69$--$88\%$ through batch size $4$, so latency-oriented small-batch serving preserves most baseline decode throughput. The cost becomes substantial in long-context, high-batch decoding: at batch size $16$, the ratio falls to $60\%$ for short inputs, $54\%$ for the 2k bucket, $49\%$ for the 8k bucket, and $44\%$ for the 32k bucket; the largest measured setting, 32k input at batch size $32$, keeps $27\%$. These matched measurements support the cost model in \S\ref{subsec:inference}: Hidden Decoding is practical for interactive or moderate-batch workloads, while long inputs and large batches expose the extra per-token computation more directly.

\section{Ablations and Probes of Intermediate Streams}
\label{sec:analysis}

Hidden Decoding relies on an unusual choice: only the final stream is trained to predict the next token, while the earlier streams receive no direct loss (\S\ref{subsec:expansion}). The matched results in \S\ref{subsec:main_results} show that Hidden Decoding improves accuracy. The accuracy gain alone leaves open what role the earlier streams play: a gain over a no-expansion baseline could come from simpler effects such as a longer expanded sequence, more prediction targets, or explicit aggregation of stream outputs. To test these possibilities, we use ablations and probes. The training-objective ablation compares final-stream supervision with all-token loss and output summation (\S\ref{subsec:sup_design}). The KV-retention analysis tests whether per-stream KV is useful, and stream probes inspect hidden-state separation and final-stream attention (\S\ref{subsec:residual}). The LM-head probes provide a vocabulary-space view of the intermediate stream states (\S\ref{subsec:dynamics}).

\subsection{Training Objective Ablation}
\label{subsec:sup_design}
To separate the effect of final-stream supervision from simpler alternatives, we train objective variants on the same 6B MoE backbone and training setup. The comparison asks whether the gain comes from leaving intermediate streams without a direct loss, or whether similar gains appear when every stream receives a prediction target or when stream outputs are explicitly aggregated. The baseline has no stream expansion. The expanded variants include \emph{all-token loss}, which applies the next-token loss to every stream; \emph{sum}, which adds the stream outputs together before prediction; and Hidden Decoding, which supervises only the final stream. The main comparison uses the same expansion factor $n{=}2$ for all expanded variants; we also report HD at $n{=}3$ to show whether the preferred objective continues to improve with more streams.

Table~\ref{tab:ablation_sup} reports the result. On language modeling, all expanded variants improve over the no-expansion baseline, but HD gives the lowest loss at the same expansion factor: $1.874$ at $n{=}2$, compared with $1.880$ for all-token loss and $1.877$ for sum. The downstream results follow the same pattern. HD at $n{=}2$ is best on all four reported benchmarks, with the clearest separation on ARC-C ($74.3$ vs.\ $71.9$ for sum and $65.8$ for all-token loss). Increasing HD to $n{=}3$ further lowers loss to $1.857$ and improves all four scores. These results favor final-stream supervision over both extra direct supervision on intermediate streams and explicit summation of stream outputs.

\begin{table}[!t]
    \centering
    \small
    \setlength{\tabcolsep}{6pt}
    \caption{\textbf{Supervision-design ablation (6B MoE).} All variants use the same backbone and training setup. At the same expansion factor $n{=}2$, applying loss to every stream (all-token loss) or summing stream outputs (sum) underperforms supervising only the final stream (HD). Loss: lower is better.}
    \label{tab:ablation_sup}
    \begin{tabular}{@{}l c cccc@{}}
    \toprule
    \textbf{Method} & \textbf{Loss}\,$\downarrow$ & MMLU & ARC-C & C-Eval & CMMLU \\
    \midrule
    Baseline                  & 1.908 & 53.4 & 68.7 & 58.4 & 61.2 \\
    All-token loss ($n{=}2$)  & 1.880 & 55.5 & 65.8 & 59.1 & 63.2 \\
    Sum ($n{=}2$)             & 1.877 & 55.6 & 71.9 & 61.0 & 63.3 \\
    \rowcolor{highlightblue}
    HD ($n{=}2$)              & 1.874 & 56.5 & 74.3 & 61.2 & 63.7 \\
    \rowcolor{highlightblue}
    HD ($n{=}3$)              & \textbf{1.857} & \textbf{58.5} & \textbf{75.7} & \textbf{62.5} & \textbf{65.9} \\
    \bottomrule
    \end{tabular}
    \end{table}

\subsection{KV Retention Ablation and Stream Probes}
\label{subsec:residual}
Hidden Decoding normally retains a separate KV cache for each stream, so later positions can use the computation produced by earlier intermediate streams. To test whether this retained per-stream KV is useful, we compare two Hidden Decoding variants that keep the same backbone, training setup, expansion factor, and Stream-Factorized Attention layout. The reference setting is the normal \emph{per-stream KV} design. The controlled variant is \emph{shared KV}, a PHD-like control \citep{wu2025efficientpretraininglengthscaling}: it keeps the final-stream prediction objective and separate stream trajectories, but removes independently retained stream KV by using one shared KV cache across streams in intra-stream layers. If the intermediate-stream KV were redundant, replacing per-stream KV with this PHD-like control should have little systematic effect.

We run this as a small supporting ablation on the 21B MoE at $n{=}2$. This setting is intentionally limited, so we use it as a qualitative check of direction. We report two Stream-Factorized Attention layouts, SF(1 full) and SF(4 full), which keep one or four full cross-stream layers. In each layout, the only intended difference between the two rows is whether KV is retained separately per stream or shared across streams.

\begin{table}[!t]
\centering
\small
\setlength{\tabcolsep}{2.4pt}
\caption{\textbf{Small KV-retention ablation (21B MoE, HD $n{=}2$).} This is a qualitative check of whether stream-specific KV matters. Accuracy columns are few-shot \texttt{loss\_acc} in percent (higher is better).}
\label{tab:kv_ablation}
\begin{tabular}{@{}c@{\hspace{0.7em}}l@{\hspace{1.0em}}cccccc@{\hspace{1.0em}}c@{}}
\toprule
\multirow{2}{*}{\makecell{\textbf{Full}\\\textbf{layers}}} & \multirow{2}{*}{\makecell[l]{\textbf{KV}\\\textbf{type}}} & \multicolumn{6}{c}{\textbf{Few-shot accuracy}} & \multirow{2}{*}{\textbf{Avg}} \\
\cmidrule(lr){3-8}
 & & ARC & EEGPQA & MMLU & MMLU-Pro & QA-MMLU & SuperGPQA & \\
\midrule
\rowcolor{highlightblue}
\multirow{2}{*}{1} & Per-stream & 91.50 & 54.17 & 74.62 & 48.83 & 80.90 & 35.36 & 64.23 \\
                  & Shared     & 91.01 & 53.19 & 74.24 & 46.60 & 81.07 & 34.63 & 63.46 \\
\midrule
\rowcolor{highlightblue}
\multirow{2}{*}{4} & Per-stream & 92.32 & 54.95 & 75.24 & 47.82 & 82.11 & 35.50 & 64.66 \\
                  & Shared     & 91.28 & 53.32 & 74.41 & 47.90 & 81.15 & 35.35 & 63.90 \\
\bottomrule
\end{tabular}
\end{table}

Because this is a small $n{=}2$ comparison, Table~\ref{tab:kv_ablation} should be read qualitatively. Shared KV lowers the average in both layouts: $64.23\!\to\!63.46$ with one full cross-stream layer and $64.66\!\to\!63.90$ with four full cross-stream layers. The per-task columns are noisy, as expected in this small comparison, but the average moves in the same direction in both layouts. These consistent average drops indicate that preserving separate KV states for different streams improves accuracy in this setting.

\begin{figure}[!t]
    \centering
    \includegraphics[width=\linewidth]{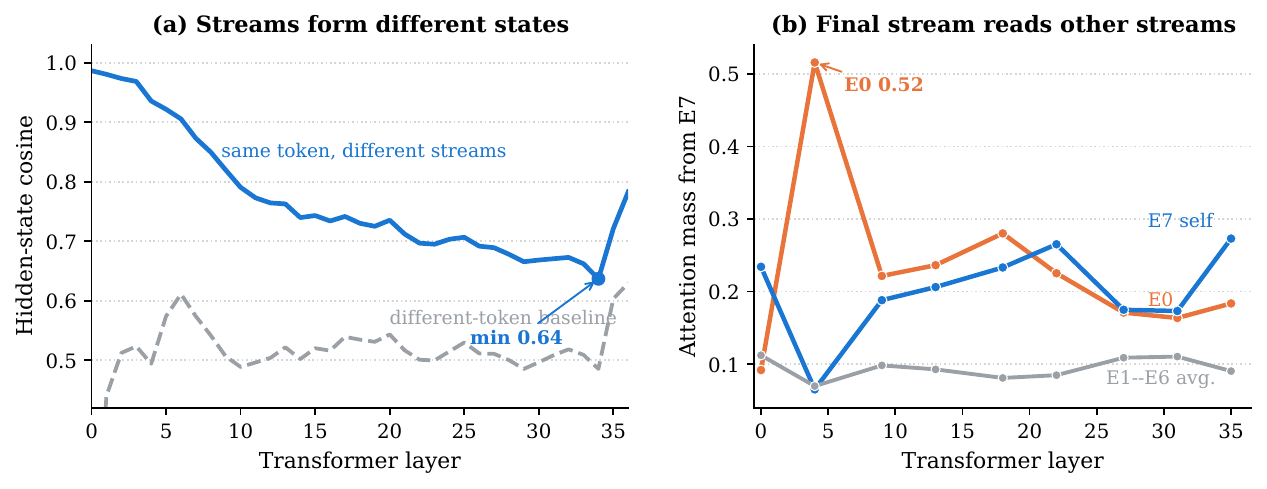}
    \caption{\textbf{Intermediate streams form states that the final stream can read (Qwen3-8B-Base + Hidden Decoding, $n{=}8$).} \textbf{(a)} Same-token streams separate inside the Transformer and partially move closer near the final layer; the dashed curve is a different-token baseline. \textbf{(b)} The final stream E7 assigns substantial attention to other streams, especially E0, showing a read path from intermediate streams to the prediction stream.}
    \label{fig:stream_mechanism}
\end{figure}

The KV ablation establishes that retaining per-stream KV affects accuracy. To connect this accuracy effect to internal computation, we inspect what the streams contain and whether the final stream reads them. We analyze the trained dense Qwen3-8B-Base Hidden Decoding model at $n{=}8$, where E7 is the final prediction stream and E0--E6 are intermediate streams. Figure~\ref{fig:stream_mechanism} gives two probes: hidden-state similarity measures whether streams form different internal states, and attention affinity measures whether E7 reads other streams.

Panel (a) shows that the streams form different internal states. Same-token streams start highly aligned, separate through the middle layers, and move closer again near the output: the mean cosine drops from $0.987$ at the input to $0.637$ in the middle layers, then rises to $0.783$ at the final layer. The curve remains above the different-token baseline, which separates stream-specific structure from generic hidden-state anisotropy. Panel (b) shows that these states are connected to the prediction stream. E7 assigns substantial attention to other streams, especially E0, whose affinity peaks at $0.52$ and is comparable to or larger than self-attention in several sampled layers. \textbf{Together, the KV ablation and the probes support the same interpretation: intermediate streams carry distinct states, and retaining their KV gives later computation access to those states.}

\subsection{LM-Head Probes in Vocabulary Space}
\label{subsec:dynamics}
To see how intermediate stream states appear at the output vocabulary level, we apply the shared LM head to each stream hidden state and inspect the resulting next-token distribution. This probe shows what token candidates are visible at different stream positions.

\begin{figure}[!t]
    \centering
    \includegraphics[width=\linewidth]{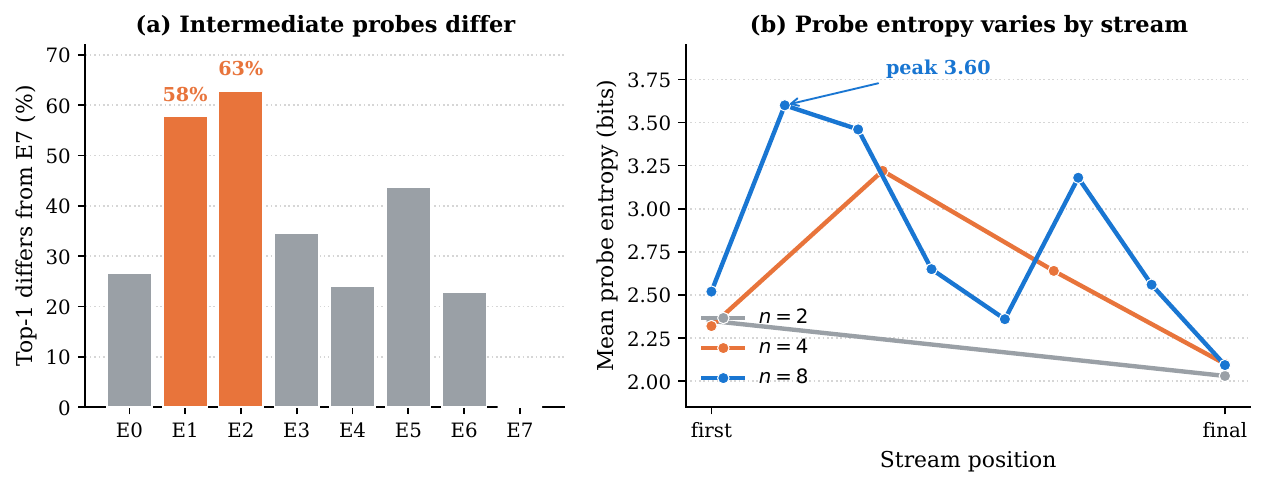}
    \caption{\textbf{LM-head probes show broader intermediate token distributions (Qwen3-8B-Base + Hidden Decoding).} We apply the shared LM head to each stream hidden state and inspect the resulting probed next-token distribution. \textbf{(a)} In the $n{=}8$ model, intermediate probes often have a top-1 token different from the final stream E7. \textbf{(b)} Intermediate probes have higher entropy than the final stream, indicating broader token-level uncertainty before the final prediction.}
    \label{fig:prediction_dynamics}
\end{figure}

Figure~\ref{fig:prediction_dynamics} gives two aggregate views. Panel (a) uses the dense Qwen3-8B-Base Hidden Decoding model at $n{=}8$ and measures how often each stream's probed top-1 token differs from the final stream E7. Intermediate probes often differ from E7, with the largest difference rate around $63\%$. Panel (b) compares mean probe entropy across HD models with $n{=}2,4,8$. The final stream has the lowest entropy in each setting; at $n{=}8$, E7 has entropy $2.09$ bits, while several intermediate streams remain above $3$ bits. \textbf{Together, the top-1 differences and higher entropy suggest that intermediate streams keep a broader token-level candidate space, while the final stream consolidates it into the prediction.}

\begin{figure}[!t]
    \centering
    \includegraphics[width=\linewidth]{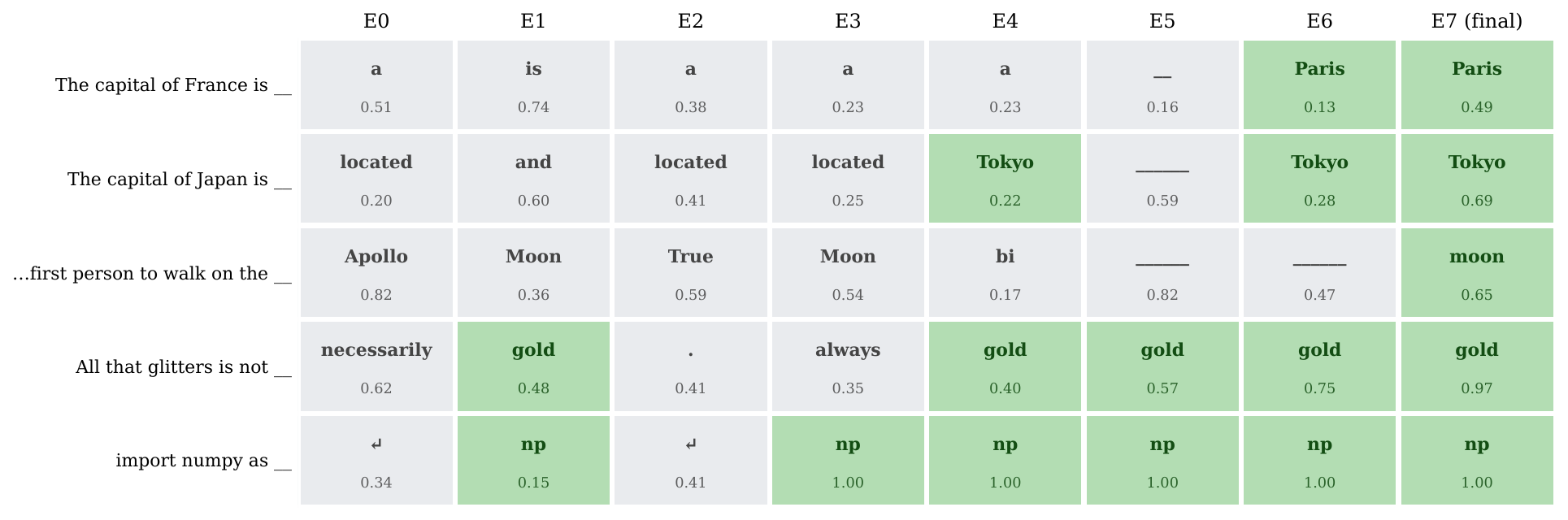}
    \caption{\textbf{Illustrative stream-probe predictions (Qwen3-8B-Base + Hidden Decoding, $n{=}8$).} For prompts with a clear answer, the figure shows each stream probe's top-1 token and its probability. Green marks probes whose top-1 matches the final stream; grey marks probes that differ. These examples serve as qualitative illustrations of the aggregate differences in Figure~\ref{fig:prediction_dynamics}.}
    \label{fig:case_study}
\end{figure}

Figure~\ref{fig:case_study} provides qualitative examples of the same pattern. Their role is illustrative: in these prompts, early or middle probes may favor template, structural, or associated tokens, such as ``a'', ``located'', or ``Apollo'', while later probes move closer to the final stream's token. These examples show the broader-candidate-space interpretation at the prompt level.

Overall, the ablations and probes support the same interpretation. The supervision-design comparison shows that final-stream supervision works better than training the intermediate streams to predict or summing their outputs. The KV and attention analyses show that stream-specific KV is useful, the streams separate inside the Transformer, and the final stream reads other streams. The LM-head probes provide an interpretability view: intermediate streams preserve a broader set of token alternatives before the final stream settles on a prediction. Together, these results support the role of earlier streams as latent computation states.

\section{Conclusion}
\label{sec:conclusion}

We introduced \textbf{Hidden Decoding}, a sequence-length scaling method for improving a fixed Transformer backbone through continued pretraining. The core idea is to give each token more internal computation by expanding it into $n$ streams, while keeping the Transformer layers unchanged. Hidden Decoding supervises only the final stream and retains the intermediate streams' KV so their computation remains available to later tokens. Stream-Factorized Attention makes this expansion trainable by keeping most layers intra-stream and limiting cross-stream mixing to a subset of layers. The ablations support these design choices: final-stream supervision outperforms direct losses on every stream or explicit summation of stream outputs, and retained per-stream KV improves accuracy in the controlled small-model checks.

The experiments show that this scaling route is useful and trainable at frontier scale. At $n{=}4$, Hidden Decoding produces WeLM-HD4-80B and WeLM-HD4-617B and improves their matched non-HD counterparts under the same early SFT-only post-training setup. The training-time measurements explain why this scale is practical: the $4\times$ expanded sequence costs $5.1\times$ on 80B and $4.4\times$ on 617B, close to linear and far below dense attention over the expanded sequence. The expansion-factor result shows that increasing $n$ improves both the dense 8B model and the 80B MoE, with 80B MMLU rising from $85.1$ to $87.5$ and Pile-test BPB falling from $0.386$ to $0.378$.

Together, these results establish sequence-length expansion as a practical fixed-backbone scaling path for frontier-scale LLMs. It does not require training a larger Transformer backbone, and it avoids the looped execution pattern that conflicts with pipeline-parallel large-model training. Instead, the extra computation appears as a longer sequence, which fits the same engineering stack used to train very large MoE models. Hidden Decoding therefore provides an effective and engineering-realistic way to keep improving strong LLMs when further backbone scaling is costly.

\bibliography{main}

\clearpage
\appendix

\section{Contributors}
\label{app:contributors}
Contributors are listed alphabetically by English name.

\begin{center}
\small
\setlength{\tabcolsep}{18pt}
\begin{tabular}{@{}lll@{}}
\toprule
Aiwei Liu & Cheng Shi & Chuhan Wu \\
Ci Lei & Di Lu & Donald He \\
Fan Zhang & Fanhao Kong & Feifei Zhang \\
Guan Wang & Haicheng Wang & Haoyu Liu \\
Houjin Yu & Jiachen Ding & Jiayi Feng \\
Jie Zhou & Jijun Chi & Jindi Shi \\
Jing Lei & Junjie Zhang & Laiyi Li \\
Le Tian & Linhao Zhang & Miao Fan \\
Sijun Zhang & Wei Jia & Weiwei Shi \\
Wenhan Li & Wentao Zhao & Wenteng Liang \\
Xiao Zhou & Xiaojin Zhou & Xihuai Wang \\
Xinyu Gao & Xuanliang Wang & Xuyang Ao \\
Yang Yu & Yangxiu You & Yinuo Zhao \\
Yufei Kuang & Yufei Wang & Yuan Liu (刘) \\
Yuan Liu (柳) & Yuwen Chen & Zhencong Tian \\
Zhongyin Zhao & Zilin Yu & Zitao Wang \\
\bottomrule
\end{tabular}
\end{center}

\section{Main WeLM Model Configurations}
\label{app:model_config}
To make the main WeLM setting reproducible, Table~\ref{tab:model_config} lists the backbone and Hidden Decoding configuration used for the 80B and 617B comparisons in \S\ref{subsec:main_results}. The non-HD and HD forms share the same Transformer backbone. Hidden Decoding changes the stream expansion and Stream-Factorized Attention layout, while leaving the Transformer widths and layer counts unchanged.

\begin{table}[H]
\centering
\small
\setlength{\tabcolsep}{3.5pt}
\caption{\textbf{Main WeLM backbone and HD configuration.} Activated parameters are per token. ``Q/KV heads'' reports query heads and grouped key--value heads. ``Experts/top-$k$'' includes one shared expert. ``KV mirror'' is the number of mirrored late layers; ``SFA'' reports intra/local/full layers.}
\label{tab:model_config}
\begin{tabular*}{\linewidth}{@{\extracolsep{\fill}}lccccccccc@{}}
\toprule
\textbf{Backbone} & \makecell{\textbf{MoE}\\\textbf{params}} & \makecell{\textbf{Activated}\\\textbf{params}} & \textbf{Layers} & \makecell{\textbf{Hidden}\\\textbf{size}} & \makecell{\textbf{Q/KV}\\\textbf{heads}} & \makecell{\textbf{Head}\\\textbf{dim.}} & \makecell{\textbf{Experts}\\\textbf{/ top-$k$}} & \makecell{\textbf{KV}\\\textbf{mirror}} & \makecell{\textbf{SFA}\\\textbf{intra/local/full}} \\
\midrule
WeLM-80B  & 80B  & 3B  & 49 & 2048 & 24/2 & 256 & $512{+}1$ / 10 & 16 & 20/23/6 \\
WeLM-617B & 617B & 23B & 94 & 4096 & 96/8 & 128 & $512{+}1$ / 10 & 30 & 69/0/25 \\
\bottomrule
\end{tabular*}
\end{table}

\section{Evaluation Configurations}
\label{app:eval}
To make the reported scores reproducible, we separate the evaluation settings into two suites. Table~\ref{tab:eval_config} gives the standard suite used for scaling studies, ablations, and base-model evaluations. Table~\ref{tab:eval_config_sft} gives the early post-training suite used for the main WeLM results in Table~\ref{tab:main_moe} and the additional 80B results in Table~\ref{tab:main_80b}.

\begin{table}[!t]
    \centering
    \small
    \setlength{\tabcolsep}{6pt}
    \caption{\textbf{Per-benchmark evaluation configurations.} Execution-based code benchmarks are scored at pass@1 with greedy decoding.}
    \label{tab:eval_config}
    \begin{tabular}{@{}llcl@{}}
    \toprule
    \textbf{Benchmark} & \textbf{Shots} & \textbf{CoT} & \textbf{Scoring} \\
    \midrule
    \rowcolor{headergray} \multicolumn{4}{@{}l}{\textit{Knowledge \& multiple-choice}} \\
    MMLU              & 5          & No  & Accuracy \\
    MMLU-Pro          & 5          & No  & Accuracy \\
    CMMLU             & 5          & No  & Accuracy \\
    C-Eval            & 5          & No  & Accuracy \\
    ARC-Challenge     & 25         & No  & Accuracy \\
    HellaSwag         & 10         & No  & Accuracy \\
    EE-GPQA           & 5          & No  & Custom evaluator \\
    SuperGPQA         & 5          & No  & Accuracy (A--K) \\
    \addlinespace
    \rowcolor{headergray} \multicolumn{4}{@{}l}{\textit{Reasoning \& math}} \\
    BBH               & 3          & Yes & EM / Accuracy \\
    GSM8K             & 4          & Yes & Numeric match \\
    MATH              & 4          & Yes & MATHEvaluator v2 \\
    MATH Modern CoT   & 4          & Yes & MATHEvaluator v2 \\
    MATH LLM Judge    & 4          & Yes & LLM-as-Judge \\
    \addlinespace
    \rowcolor{headergray} \multicolumn{4}{@{}l}{\textit{Factual QA}} \\
    SimpleQA          & 5          & No  & Substring match \\
    Chinese SimpleQA  & 10         & No  & LLM Judge \\
    AA-OmniScience    & 10         & No  & LLM Judge \\
    \addlinespace
    \rowcolor{headergray} \multicolumn{4}{@{}l}{\textit{Code}} \\
    CRUXEval          & 1 / 2      & Yes & Execution (pass@1) \\
    EvalPlus          & 1          & No  & Execution (pass@1) \\
    MultiPL-E         & 1          & No  & Execution (pass@1) \\
    \bottomrule
    \end{tabular}
    \end{table}

\begin{table}[!t]
\centering
\small
\setlength{\tabcolsep}{6pt}
\caption{\textbf{Evaluation settings for the early post-training suite (Tables~\ref{tab:main_moe} and~\ref{tab:main_80b}).} $N$ is the number of independent runs per question; scores are averaged over the $N$ runs. LLM-judge benchmarks use a fixed judge model and temperature. $^{\ddagger}$HMMT is the weighted average over the Feb.\ 2025, Nov.\ 2025, and Feb.\ 2026 sets. $^{\dagger}$External evaluations.}
\label{tab:eval_config_sft}
\begin{tabular}{@{}l c l@{}}
\toprule
\textbf{Benchmark} & \textbf{$N$} & \textbf{Scoring} \\
\midrule
GPQA Diamond      & 4  & Accuracy (multiple choice) \\
HMMT$^{\ddagger}$ & 4  & Weighted average; LLM judge (answer equiv.) \\
IMO-AnswerBench   & 1  & LLM judge (answer equiv.) \\
MathArena Apex    & 32 & LLM judge (HLE-style) \\
FrontierMath      & 32 & LLM judge (HLE-style) \\
PolyMath          & 1  & LLM judge (HLE-style) \\
PHYBench          & 4  & LLM judge (HLE-style) \\
CritPt            & 8  & Two-call, CritPt API \\
HLE               & 1  & LLM judge (HLE-style) \\
MMMLU             & 1  & LLM judge (HLE-style) \\
AA-OmniScience    & 2  & LLM judge (4-way) \\
AA-LCR            & 3  & LLM judge (equality) \\
SciCode           & 1  & Execution (sub-step rate) \\
LiveCodeBench v6  & 2  & Execution (pass@1) \\
IFBench           & 2  & Programmatic checks (pass@1) \\
ARC-AGI-2         & 1  & Exact grid match (pass@2) \\
Terminal-Bench 2  & 3  & Agent + verifier (resolved rate) \\
GDPval$^{\dagger}$            & -- & Win rate \\
$\tau^2$-Bench$^{\dagger}$    & -- & Reward \\
\bottomrule
\end{tabular}
\end{table}

Table~\ref{tab:eval_config} reports the few-shot count, chain-of-thought (CoT) setting, and scoring method for each benchmark in the standard suite. For SuperGPQA, EE-GPQA, Chinese SimpleQA, and AA-OmniScience, the few-shot exemplars are pre-formatted as a fixed prefix in the data rather than retrieved dynamically; the same exemplars are therefore shared across all examples of those benchmarks. The three MATH variants share the same dataset and differ only in scoring: rule-based equivalence (v2), the same scorer with extra Chinese stopping tokens (modern CoT), and an LLM judge. This lets us separate scoring effects from model effects.

The standard suite spans four families. Knowledge and multiple-choice benchmarks are scored by exact-match accuracy on the predicted option letter, without chain-of-thought. Reasoning and math benchmarks use chain-of-thought prompting and then extract the final answer. Factual-QA benchmarks ask for short answers, scored either by normalized substring matching (SimpleQA) or by an LLM judge (Chinese SimpleQA, AA-OmniScience). Code benchmarks execute the generated programs against test cases and report pass@1 under greedy decoding. The LLM-judge benchmarks use a fixed judge model and temperature throughout, so scores stay comparable across our runs.

Table~\ref{tab:eval_config_sft} lists the settings for the early post-training suite. Each question is run $N$ times and the per-question scores are averaged; $N$ is as large as $32$ for the small, high-variance math sets (MathArena Apex, FrontierMath). Generation is free-form with a generous token budget (up to $160$k tokens per response), so the models can use long chains of thought. The scoring styles are as follows.

Most math, science, and knowledge benchmarks are open-ended: the model writes a full solution, and an LLM judge compares its final answer to the gold answer for equivalence, following the protocol of Humanity's Last Exam~\citep{phan2025hle} (we refer to this as HLE-style judging). This covers FrontierMath, MathArena Apex, PHYBench, PolyMath, MMMLU, and HLE itself; the math benchmarks use Qwen3.5-397B-A17B as a fixed judge model. GPQA Diamond is the exception---it is multiple-choice, scored by whether the selected option matches the correct label. AA-OmniScience uses a four-way judge (correct / incorrect / partial / not-attempted) and reports accuracy, and AA-LCR uses a general equality judge over long-document questions.

The remaining benchmarks are scored by execution or by task-specific verifiers rather than a judge. SciCode generates code for each sub-step of a problem and runs it against reference assertions, scoring the fraction of sub-steps that pass; LiveCodeBench~v6 and IFBench run the generated programs (or instruction-checking code) and report pass@1. ARC-AGI-2 parses the predicted grid from the response and requires an exact match, reporting task-level pass@2. Terminal-Bench~2 is agentic: for each task the model drives an agent inside a containerized environment, and an environment verifier returns a reward, which we report as the resolved rate. CritPt makes two model calls per question---first solving the physics problem, then filling in a code template---and submits the result to the CritPt scoring API. Across all LLM-judge benchmarks the judge model and temperature are fixed throughout, so scores stay comparable across our runs.

\section{Early SFT Hyperparameters}
\label{app:sft_recipe}
To make the early post-training comparison in Table~\ref{tab:main_moe} reproducible, we use the same SFT hyperparameters for each matched pair of non-HD and HD models. This supervised-only stage tests whether the gains from continued pretraining persist after the same lightweight adaptation step.

\begin{table}[!t]
\centering
\small
\setlength{\tabcolsep}{6pt}
\caption{\textbf{Early SFT hyperparameters for the matched WeLM comparisons.} The same recipe is used for each matched pair of non-HD and HD models in Table~\ref{tab:main_moe}; no reinforcement learning stage is applied.}
\label{tab:sft_recipe}
\begin{tabular}{@{}ll@{}}
\toprule
\textbf{Item} & \textbf{Setting} \\
\midrule
Optimizer & Adam + Muon \\
Adam parameters & $\beta_1{=}0.9$, $\beta_2{=}0.95$, $\epsilon{=}10^{-8}$ \\
Muon parameters & Momentum $0.95$, Nesterov enabled, $5$ Newton--Schulz steps \\
Learning rate & $2.0{\times}10^{-5}$ \\
Schedule & Cosine decay to $0$, with $56$ warmup iterations and $1125$ decay iterations \\
Regularization & Weight decay $0.1$; gradient clipping $1.0$ \\
Precision & bfloat16 \\
Batching & Global token batch $4{,}194{,}304$; micro-batch token budget $262{,}144$ \\
Context extension & YaRN scaling factor $8$ from a $32$k base context; rotary base $500{,}000$ \\
Adaptation scope & LoRA disabled; supervised fine-tuning only, with no RL \\
\bottomrule
\end{tabular}
\end{table}

\section{Additional Benchmarks for WeLM-HD4-80B}
\label{app:80b_extra}
To check whether the 80B gains extend beyond the nine benchmarks shared across scales in Table~\ref{tab:main_moe}, we evaluate WeLM-HD4-80B on a broader suite covering agentic, code, instruction-following, and long-context tasks. Table~\ref{tab:main_80b} compares WeLM-80B and WeLM-HD4-80B under the same early SFT-only post-training setup as \S\ref{subsec:main_results}. WeLM-HD4-80B improves eight of the ten benchmarks, with the largest gains on Terminal-Bench 2 ($44.9\!\to\!58.4$) and ARC-AGI-2 ($6.9\!\to\!11.6$).

\begin{table}[!t]
\centering
\small
\setlength{\tabcolsep}{6pt}
\caption{\textbf{Additional benchmarks for WeLM-HD4-80B.} The highlighted column is WeLM-HD4-80B, compared against the matched non-HD WeLM-80B under the same early SFT-only post-training recipe as Table~\ref{tab:main_moe}. $^{\dagger}$External evaluations outside our harness.}
\label{tab:main_80b}
\begin{tabular}{@{}l c >{\columncolor{highlightblue}}c@{}}
\toprule
\textbf{Benchmark} & \makecell{\textbf{WeLM}\\\textbf{80B}} & \makecell{\textbf{WeLM-HD4}\\\textbf{80B}} \\
\midrule
Terminal-Bench 2  & 44.9 & 58.4 \\
ARC-AGI-2         & 6.9  & 11.6 \\
$\tau^2$-Bench$^{\dagger}$   & 75.4 & 78.9 \\
GDPval$^{\dagger}$           & 39.6 & 42.9 \\
PolyMath          & 59.4 & 62.3 \\
IFBench           & 70.7 & 73.0 \\
LiveCodeBench v6  & 83.7 & 85.1 \\
CritPt            & 3.7  & 4.9  \\
AA-OmniScience    & 27.7 & 27.6 \\
AA-LCR            & 69.3 & 69.3 \\
\bottomrule
\end{tabular}
\end{table}

\section{Stream-Factorized Attention Layouts}
\label{app:sfa}
To make the attention-cost analysis in \S\ref{sec:cost} and the ablation in Table~\ref{tab:ablation_attn} reproducible, Table~\ref{tab:sfa_layout} lists the per-model layer composition used for Stream-Factorized Attention. Intra-stream layers attend only within a stream. Cross-stream layers mix information across streams and follow the base model's attention pattern: ``local'' denotes sliding-window cross-stream attention, and ``full'' denotes full cross-stream attention.

The three layer types differ sharply in cost over the expanded length-$nL$ sequence: an intra-stream layer costs $O(nL^2)$, a local cross-stream layer costs $O(nLw)$ for window size $w$, and a full cross-stream layer costs $O(n^2L^2)$. Full cross-stream layers are the only layers with quadratic growth in $n$, so the layouts keep their number small while still letting every token mix across streams. For base models that already interleave sliding-window and full attention (21B and 80B), the local cross-stream layers reuse the base model's sliding-window layers; the 617B base uses full attention throughout, so its cross-stream layers are all full ($25$ of $94$), with the rest converted to intra-stream.

\begin{table}[!t]
\centering
\small
\setlength{\tabcolsep}{6pt}
\caption{\textbf{Per-model Stream-Factorized Attention layouts.} ``Local'' denotes sliding-window cross-stream layers (used when the base model has SWA); ``Full'' denotes full-attention cross-stream layers. The three 21B rows correspond to the configurations ablated in Table~\ref{tab:ablation_attn}.}
\label{tab:sfa_layout}
\begin{tabular}{@{}lcccc@{}}
\toprule
\textbf{Model} & \textbf{Layers} & \textbf{Intra} & \textbf{Local} & \textbf{Full} \\
\midrule
21B, all-full       & 27 & 0  & 0  & 27 \\
21B, SF ($4$ full)  & 27 & 10 & 13 & 4  \\
21B, SF ($1$ full)  & 27 & 13 & 13 & 1  \\
80B                 & 49 & 20 & 23 & 6  \\
617B                & 94 & 69 & 0  & 25 \\
\bottomrule
\end{tabular}
\end{table}

\section{Progressive Expansion: Full Results}
\label{app:progressive}
To show the full expansion-factor trajectory behind \S\ref{subsec:progressive_exp}, Tables~\ref{tab:prog_80b} and~\ref{tab:prog_8b} report results on the 80B MoE that yields WeLM-HD4-80B at $n{=}4$ and on the dense Qwen3-8B-Base. Only the embedding tables grow with $n$; the backbone and the active computation per token are held fixed. The best value in each row is in bold.

We introduce the higher factors during continued pretraining at scheduled token counts, letting each factor converge before the next. On the 80B MoE, we expand to $n{=}2$, $4$, and $8$ after $0$, $503$B, and $906$B tokens of the 32k continuation phase. The tables report two complementary signals: a steady reduction in language-modeling loss (Pile-test BPB on the 80B model, from $0.386$ to $0.378$) and consistent downstream gains, with $n{=}8$ best on almost every benchmark. The few exceptions are non-monotonic only at intermediate factors (e.g., MBPP+ on 80B dips at $n{=}2$ before recovering), while the $n{=}8$ model remains the strongest overall.

\begin{table}[!t]
    \centering
    \small
    \setlength{\tabcolsep}{6pt}
\caption{\textbf{Progressive expansion on the 80B MoE toward WeLM-HD4-80B.} Active parameters per token are fixed at $3$B; only embedding parameters grow with $n$. BPB: lower is better.}
    \label{tab:prog_80b}
    \begin{tabular}{@{}l c c >{\columncolor{highlightblue}}c >{\columncolor{highlightblue}}c >{\columncolor{highlightblue}}c@{}}
    \toprule
    \textbf{Benchmark} & \textbf{Shots} & \textbf{Base} & $n{=}2$ & $n{=}4$ & $n{=}8$ \\
    \midrule
    Embedding params  & -- & 6.1B  & 12.1B & 24.2B & 48.4B \\
    Training tokens   & -- & 1.37T & 503B  & 906B  & 1.01T \\
    \midrule
    Pile-test (BPB)\,$\downarrow$ & --  & 0.386 & 0.387 & 0.382 & \textbf{0.378} \\
    BBH (EM)           & 3   & 87.5 & 88.3 & 90.0 & \textbf{90.6} \\
    MMLU (EM)          & 5   & 85.1 & 85.0 & 86.7 & \textbf{87.5} \\
    C-Eval (EM)        & 5   & 88.8 & 88.9 & \textbf{89.9} & 89.5 \\
    SimpleQA           & 5   & 16.7 & 15.1 & 17.2 & \textbf{18.4} \\
    Chinese SimpleQA   & 5   & 60.7 & 62.2 & 63.8 & \textbf{64.7} \\
    HumanEval+         & 1   & 61.0 & 61.0 & 59.1 & \textbf{62.2} \\
    MBPP+              & 1   & 67.7 & 64.4 & 70.2 & \textbf{71.2} \\
    MATH               & 4   & 60.4 & 58.4 & 71.4 & \textbf{71.7} \\
    \bottomrule
    \end{tabular}
    \end{table}

\begin{table}[!t]
\centering
\small
\setlength{\tabcolsep}{6pt}
\caption{\textbf{Progressive expansion on the dense Qwen3-8B-Base.} Total parameters are fixed at $8$B; only embedding parameters grow with $n$.}
\label{tab:prog_8b}
\begin{tabular}{@{}l c c >{\columncolor{highlightblue}}c >{\columncolor{highlightblue}}c >{\columncolor{highlightblue}}c@{}}
\toprule
\textbf{Benchmark} & \textbf{Shots} & \textbf{Base} & $n{=}2$ & $n{=}4$ & $n{=}8$ \\
\midrule
Embedding params  & -- & 1.2B & 1.9B & 3.1B & 5.6B \\
Training tokens   & -- & 180B & 75B  & 150B & 187B \\
\midrule
BBH (EM)          & 3   & 78.8 & 81.3 & 83.0 & \textbf{83.9} \\
MMLU (EM)         & 5   & 79.8 & 80.9 & 81.9 & \textbf{82.2} \\
ARC-C             & 25  & 93.9 & 94.3 & 94.4 & \textbf{94.7} \\
HellaSwag         & 10  & 79.7 & 83.1 & 85.0 & \textbf{85.3} \\
GSM8K             & 4   & 92.5 & 93.3 & 93.9 & \textbf{94.6} \\
MATH              & 4   & 56.0 & 58.2 & 60.0 & \textbf{61.1} \\
MBPP+             & 1   & 66.7 & 69.4 & 68.7 & \textbf{69.4} \\
\bottomrule
\end{tabular}
\end{table}

\section{CPT Startup Loss under Attention Compositions}
\label{app:cpt_startup_loss}

To test whether using more intra-stream layers gives a smoother starting point for continued pretraining, we compare three $n{=}4$ attention compositions on a separate 617B-scale checkpoint. All runs start from the same step-16000 checkpoint before HD expansion and enable the expanded streams at the next step. The model, data, training schedule, and expansion factor are matched; the variants differ only in attention composition: no intra-stream layers, a 1:1 cross-/intra-stream layout, and a 1:3 cross-/intra-stream layout.

\begin{figure}[!t]
    \centering
    \includegraphics[width=\linewidth]{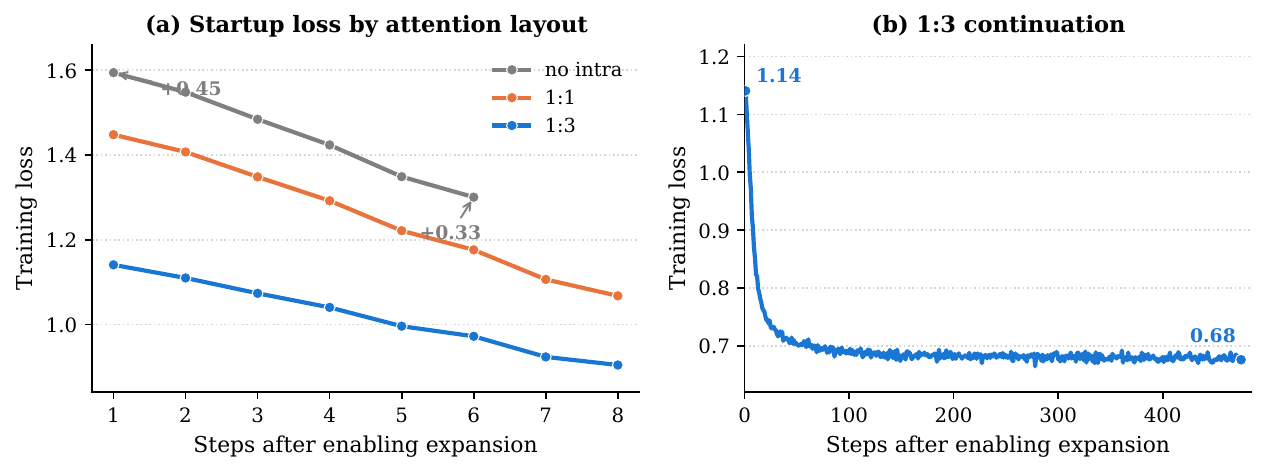}
    \caption{\textbf{CPT startup loss under different attention compositions.} All runs enable $n{=}4$ expansion from the same 617B-scale step-16000 checkpoint. \textbf{(a)} The startup window compares no intra-stream layers, a 1:1 cross-/intra-stream layout, and a 1:3 cross-/intra-stream layout. The run without intra-stream layers and the 1:1 run stop after 6 and 8 steps, so the comparison is restricted to early startup behavior. \textbf{(b)} The 1:3 layout continues training for 475 steps.}
    \label{fig:startup_loss}
\end{figure}

Figure~\ref{fig:startup_loss} shows the loss immediately after the expansion is enabled. At the first expanded step, the losses are $1.59$ for no intra-stream layers, $1.45$ for the 1:1 layout, and $1.14$ for the 1:3 layout. The layout with no intra-stream layers is higher than the 1:3 layout by $+0.45$ at step 1 and still by $+0.33$ at step 6. The ordering no intra-stream $>$ 1:1 $>$ 1:3 holds throughout the shared six-step window.

The run without intra-stream layers and the 1:1 run stop after 6 and 8 steps, so this check is limited to startup behavior rather than final convergence. The 1:3 layout continues training for 475 steps. These results support the design role of intra-stream layers: they reduce attention cost and also reduce the startup loss increase when the expanded model starts CPT from a checkpoint before HD expansion.

\section{Base-Model Results with and without Hidden Decoding}
\label{app:pretrain}
To check whether the gains appear before post-training, Table~\ref{tab:main_pretrain} compares each non-HD Base model against its $n{=}4$ Hidden Decoding Base model.

\begin{table}[!htbp]
    \centering
    \small
    \setlength{\tabcolsep}{6pt}
    \caption{\textbf{WeLM Base models with and without Hidden Decoding.}}
    \label{tab:main_pretrain}
    \begin{tabular}{@{}l cc cc@{}}
    \toprule
     & \multicolumn{2}{c}{\textbf{80B MoE}} & \multicolumn{2}{c}{\textbf{617B MoE}} \\
    \cmidrule(lr){2-3}\cmidrule(lr){4-5}
    \textbf{Benchmark} & \makecell{\textbf{WeLM}\\\textbf{80B-Base}} & \cellcolor{highlightblue}\makecell{\textbf{WeLM-HD4}\\\textbf{80B-Base}} & \makecell{\textbf{WeLM}\\\textbf{617B-Base}} & \cellcolor{highlightblue}\makecell{\textbf{WeLM-HD4}\\\textbf{617B-Base}} \\
    \midrule
    MMLU              & 88.0 & \cellcolor{highlightblue}88.9 & 89.85 & \cellcolor{highlightblue}90.22 \\
    MMLU-Pro          & 68.4 & \cellcolor{highlightblue}71.3 & 72.26 & \cellcolor{highlightblue}73.92 \\
    C-Eval            & 91.3 & \cellcolor{highlightblue}91.4 & 91.9 & \cellcolor{highlightblue}92.5 \\
    SuperGPQA         & 50.5 & \cellcolor{highlightblue}52.6 & 55.33 & \cellcolor{highlightblue}57.05 \\
    BBH               & 90.7 & \cellcolor{highlightblue}91.9 & 92.41 & \cellcolor{highlightblue}93.07 \\
    MATH              & 63.6 & \cellcolor{highlightblue}62.6 & 70.92 & \cellcolor{highlightblue}70.82 \\
    SimpleQA          & 14.9 & \cellcolor{highlightblue}15.3 & 41.47 & \cellcolor{highlightblue}41.75 \\
    Chinese SimpleQA  & 56.0 & \cellcolor{highlightblue}57.4 & 78.13 & \cellcolor{highlightblue}78.8 \\
    AA-OmniScience    & 22.2 & \cellcolor{highlightblue}24.9 & 36.78 & \cellcolor{highlightblue}37.12 \\
    HumanEval+        & 70.1 & \cellcolor{highlightblue}72.0 & 76.2 & \cellcolor{highlightblue}76.2 \\
    MBPP+             & 70.9 & \cellcolor{highlightblue}71.4 & 71.4 & \cellcolor{highlightblue}70.9 \\
    \midrule
    \textbf{Average} & 62.42 & \cellcolor{highlightblue}\textbf{63.61} & 70.60 & \cellcolor{highlightblue}\textbf{71.12} \\
    \bottomrule
    \end{tabular}
    \end{table}

Hidden Decoding improves both scales before post-training. The average rises from $62.42$ to $63.61$ on 80B and from $70.60$ to $71.12$ on 617B. The larger gains appear on harder reasoning and knowledge tasks such as SuperGPQA and MMLU-Pro, and Table~\ref{tab:main_moe} shows that they persist after the same lightweight post-training recipe.

\end{CJK*}
\end{document}